\documentclass[lettersize,journal,compsoc]{IEEEtran}
\usepackage{amsmath,amsfonts}
\usepackage{algorithm}
\usepackage{array}
\usepackage{subfigure} 
\usepackage{textcomp}
\usepackage[noend]{algpseudocode} 
\usepackage{stfloats}
\usepackage{url}
\usepackage{verbatim}
\usepackage{graphicx}
\usepackage{cite}
\usepackage{multirow}
\usepackage{tabulary}
\usepackage[table]{xcolor}
\usepackage{amssymb}
\usepackage{booktabs}
\usepackage{color} 
\usepackage{xcolor}
\usepackage{bbding}
\usepackage{float}  %设置图片浮动位置的宏包
\usepackage{enumitem} 
\usepackage{enumerate}
\usepackage{pifont}
\usepackage{cleveref}
\usepackage{amsmath}
\usepackage{amsthm}
\usepackage{amssymb}
\newtheorem{theorem}{Theorem}
\newtheorem{assumption}{Assumption}
\begin{document}

\title{Multi-level Personalized Federated Learning on Heterogeneous and Long-Tailed Data}
% CPFL short for 
% \title{Cluster-based / Multi-level Personalized Federated Learning on non-i.i.d. and Long-Tailed data} ?

\author{Rongyu Zhang, Yun Chen, Chenrui Wu, Fangxin Wang, and Bo Li

\thanks{This paper is an extended version of \cite{zhang2023cluster}, in proceedings of IEEE ICME 2023.}
\thanks{(Corresponding author: Fangxin Wang.)}
\thanks{Rongyu Zhang, Yun Chen(equal contribution), Chenrui Wu(equal contribution) are with the Future Network of Intelligence Institute (FNii) and the School of Science and Engineering (SSE), The Chinese University of Hong Kong, Shenzhen (e-mail: \{rongyuzhang, yunchen1, chenruiwu \}@link.cuhk.edu.cn).}
\thanks{Fangxin Wang is with the School of Science and Engineering (SSE) and the Future Network of Intelligence Institute (FNii), The Chinese University of Hong Kong, Shenzhen, with the Guangdong Provincial Key Laboratory of Future Networks of Intelligence (e-mail: wangfangxin@cuhk.edu.cn).}
\thanks{Bo Li is the Chair Professor in the Department of Computer Science and Engineering at Hong Kong University of Science and Technology (email: bli@ust.hk).}
}

% \title{Multi-level Personalized Federated Learning on Heterogeneous and Long-Tailed data}

% \author{Rongyu Zhang, Yun Chen\thanks{Rongyu Zhang, Yun Chen (equal contributions), Chenrui Wu, and Fangxin Wang (corresponding author) are with The Chinese University of Hong Kong, Shenzhen, China. Jiangchuan Liu is with Simon Fraser University, Canda.}, Chenrui Wu, Fangxin Wang,~\IEEEmembership{Member,~IEEE,} Jiangchuan Liu,~\IEEEmembership{Fellow,~IEEE,}}

% \markboth{IEEE TRANSACTIONS ON MOBILE COMPUTING, VOL., NO., 2023}%

\IEEEtitleabstractindextext{
\begin{abstract} 
%Federated learning (FL) empowers a privacy-preserving distributed learning framework where models are trained at each client and aggregated to a central server without data transmission. However, in realistic FL scenarios, data distribution in mobile devices is often non-i.i.d. and class distribution is severely long-tailed, which can easily lead to overfitting. This is because the trained local models are prone to get stuck into the local optimum after model aggregation with biased parameters. In this paper, we study how heterogeneous data causes model bias and further propose a novel personalized federated framework named Multi-level Personalized Federated Learning \texttt{(MuPFL)} with three major modules: Biased Activation Value Dropout (\textit{BAVD}) extracts the important activation to prevent overfitting and accelerate model training, Adaptive Cluster-based Model Update (\textit{ACMU}) smooths the biased local model parameters for more effective global aggregation, and Prior Knowledge-assisted Classifier Fine-tuning (\textit{PKCF}) is applied on the classifier with shared global prior knowledge to enhance its classification ability and generate personalized models based on the local long-tailed data. Extensive experimental results on multiple real-world datasets on both image classification and semantic segmentation tasks demonstrate that our proposed method outperforms other state-of-the-art baselines even when data is highly non-i.i.d. and long-tailed.
Federated learning (FL) offers a privacy-centric distributed learning framework, enabling model training on individual clients and central aggregation without necessitating data exchange. Nonetheless, FL implementations often suffer from non-i.i.d. and long-tailed class distributions across mobile applications, e.g., autonomous vehicles, which leads models to overfitting as local training may converge to sub-optimal.
In our study, we explore the impact of data heterogeneity on model bias and introduce an innovative personalized FL framework, Multi-level Personalized Federated Learning (\texttt{MuPFL}), which leverages the hierarchical architecture of FL to fully harness computational resources at various levels. This framework integrates three pivotal modules: Biased Activation Value Dropout (BAVD) to mitigate overfitting and accelerate training; Adaptive Cluster-based Model Update (ACMU) to refine local models ensuring coherent global aggregation; and Prior Knowledge-assisted Classifier Fine-tuning (PKCF) to bolster classification and personalize models in accord with skewed local data with shared knowledge. Extensive experiments on diverse real-world datasets for image classification and semantic segmentation validate that \texttt{MuPFL} consistently outperforms state-of-the-art baselines, even under extreme non-i.i.d. and long-tail conditions, which enhances accuracy by as much as 7.39\% and accelerates training by up to 80\% at most, marking significant advancements in both efficiency and effectiveness.
\end{abstract} 

\begin{IEEEkeywords}
Federated learning, Long-tailed learning, Dropout, Clustering, Personalization, Autonomous driving
\end{IEEEkeywords}
}
\maketitle

% \footnote{Portions of this work were presented at the International Conference on Multimedia and Expo (ICME) in 2023, \textit{Cluster-driven GNN-based Federated Recommendation with Biased Message Dropout (CdFed)}, which updates the local and global models from multi-level including local level, interface level, and global level according to our empirical observation to tackle the long-tailed challenges and obtain a personalized model for each client. Compared with \textit{CdFed}, we shifted the research background from the recommendation system to image classification in the wider visual field. At the same time, we went deep into the non-i.i.d. problem to explore and solve more challenging long-tail problems.}

\section{Introduction}

\IEEEPARstart{T}{he} merge of 5G and mobile edge computing~\cite{li2018energy,zhang2023repcam,zhou2023mec,zhang2021computation} (MEC) has sparked a transformative shift in the Internet of Things~\cite{jiang2020energy,ding2018beef,kumar2019target} (IoT) ecosystem, paving the way for an era of distributed collective intelligence across multimedia platforms. This shift is marked by the widespread deployment of IoT devices—from smart home gadgets to sophisticated autonomous driving systems—each producing voluminous data that enriches the network's intelligence, significantly impacting everyday life with enhanced connectivity and smarter solutions.

Conventionally, distributed machine learning has anchored by the paradigm of \textit{data aggregation}, which involves the consolidation of local datasets at a centralized cloud server for subsequent model training \cite{pan2023cloud}. Yet, this method is fraught with privacy risks and the potential for data leakage, a critical issue in domains like finance and healthcare where confidentiality is crucial. The challenge is exacerbated in the context of autonomous vehicles, where transmitting vast quantities of image and video data from on-board cameras to a central server can consume extensive network bandwidth and introduce latency incompatible with the real-time decision-making demands of autonomous navigation. Federated learning (FL), as delineated in seminal works\cite{mcmahan2017communication,sahu2018convergence,wang2020tackling,karimireddy2020scaffold,zhang2023unimodal}, offers a compelling alternative through \textit{model aggregation}. This technique\cite{yang2019federated, li2020survey,feddyn,dai2023tackling} facilitates collaborative training across clients while preserving data confidentiality.

\begin{figure}[t]
    \centering
    \includegraphics[width=0.48\textwidth]{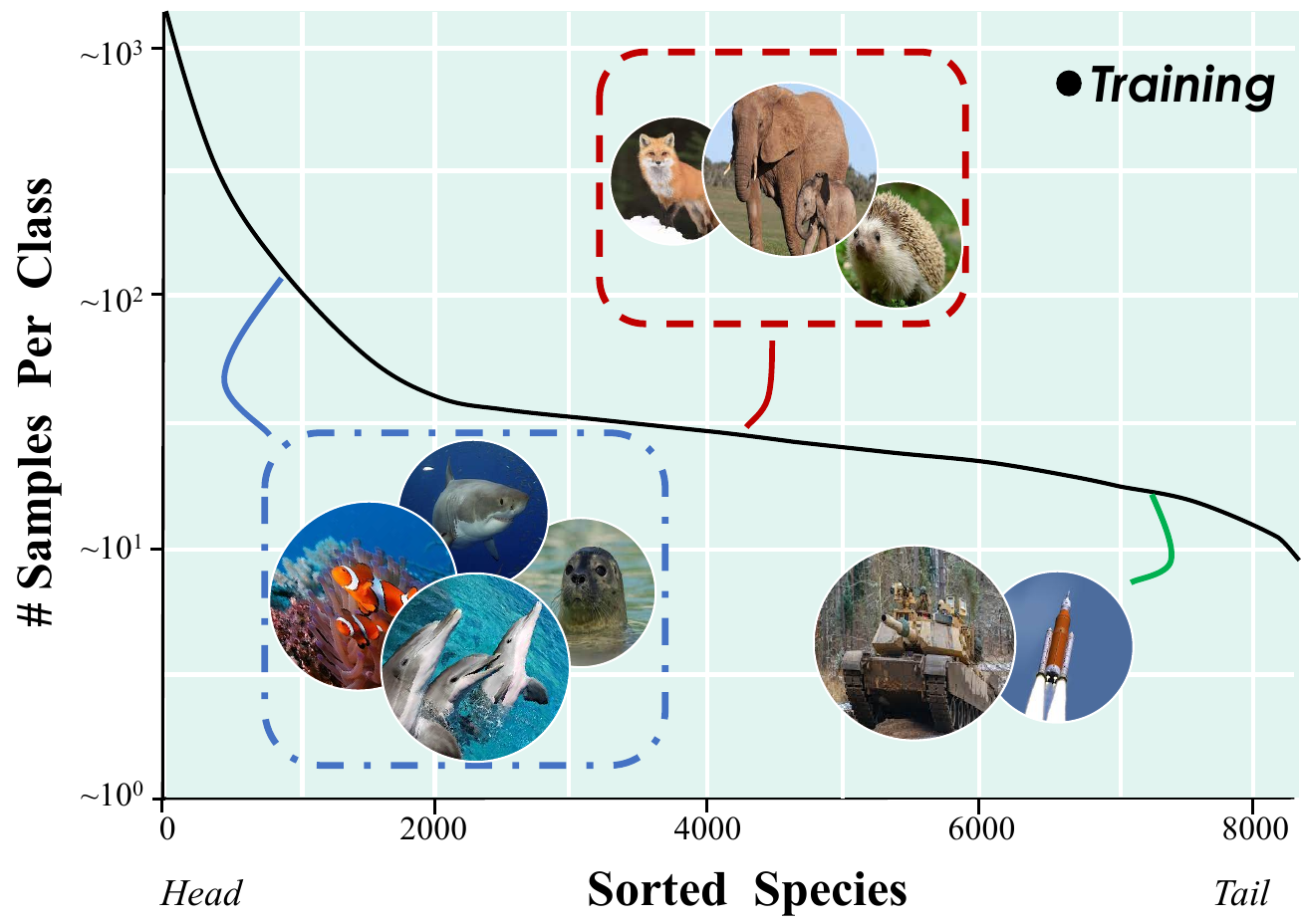}
    \caption{Exhibition of a long-tailed training set data distribution in the iNaturalist 2018 dataset\cite{2018The} with over 8,000 species, making it a challenging dataset for classification. Some species like sharks and iNaturalist 2018 dataset have many images, while others have very few.}
    \label{long-tailed}
\end{figure}

\begin{figure}[t]
    \centering
    \includegraphics[width=0.48\textwidth]{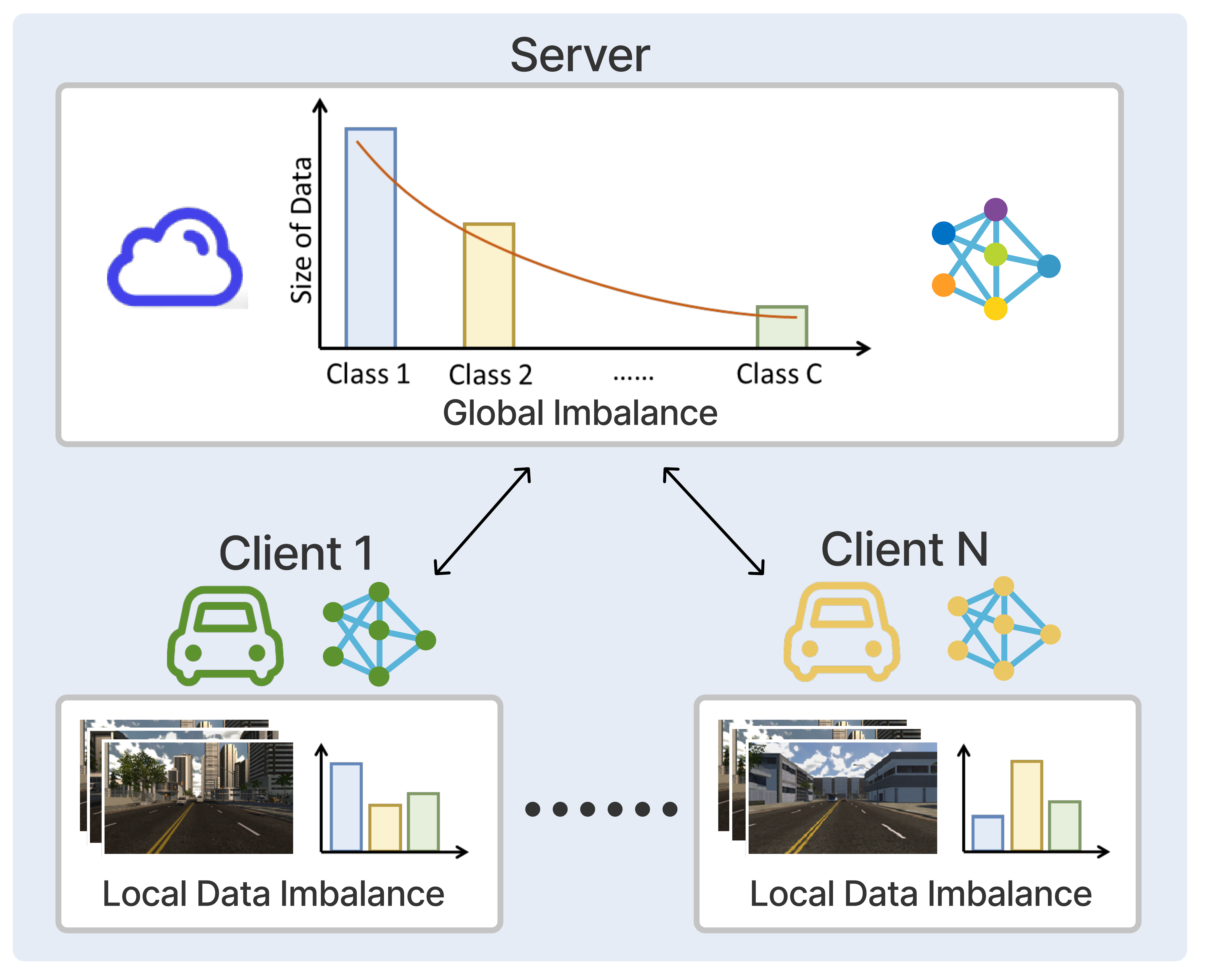}
    \caption{The illustration of long-tailed data in personalized federated learning scenarios for autonomous driving.}
    \label{fig:framwork-0}
\end{figure}

Existing works~\cite{zhao2018federated, hsieh2020non,wu2023fedab,wu2023learning} have endeavored to further enhance the model performance within FL. However, these efforts often assumed an independent and identically class distribution (i.i.d)—a condition rarely met in practical situations, particularly in autonomous driving applications where data distribution is highly heterogeneous across different locations. In the case of autonomous vehicles, on-board sensors capture a plethora of instances, ranging from traffic signs to public transit with varying frequency and distribution. This results in a more challenging long-tailed distribution than conventional non-i.i.d scenarios as shown in \cref{long-tailed} and \cref{fig:framwork-0}. Such class imbalance is prone to biased global model aggregation, adversely affecting performance for FL system on tail classes. In this work, we identify three principal challenges for long-tailed federated learning, Challenge \ding{202}, the mitigation of overfitting within tail classes and enhancement of training efficiency; Challenge \ding{203}, the avoidance of biased model global aggregation due to non-i.i.d. and long-tail distributions; and Challenge \ding{204}, the improvement of classification and segmentation performance for tail classes within local models, which is a requisite for mobile applications.

In response to the outlined challenges, Personalized Federated Learning (PFL) has been prevailing \cite{smith2017federated, zantedeschi2020fully, zhang2023optimizing, pmlr-v139-collins21a,fedrod} to customize the local model for each local client. These approaches range from the creation of task similarity graphs to the adoption of personalized regularization components, all are meticulously crafted to boost the precision of FL models. Despite their effectiveness, these solutions generally neglect the exploration in the local training stage to extract the data distribution in the most direct manner. Moreover, they often incur substantial computational burdens on the client side, which pose significant misalignment with the operational realities of mobile computing environments where computational resources are inherently limited.

In this context, we present an innovative and computationally efficient framework termed \textit{Multi-level Personalized Federated Learning} (\texttt{MuPFL}) that capitalizes on the intrinsic benefits of a hierarchical federated learning architecture, anchored by a central server with unlimited resources. \texttt{MuPFL} is ingeniously crafted to utilize the robust processing capabilities of the server, thereby alleviating the computational strain on client devices and facilitating a more sustainable and personalized federated learning paradigm. It encompasses three innovative components corresponding to three different levels. Module \ding{202} in local clients: Biased Activation Value Dropout (BAVD) to counteract overfitting and accelerate training by updating activation maps personalized to each local model's epoch; Module \ding{203} in middle interface: Adaptive Cluster-based Model Update (ACMU) an intermediate step between server and client, employing cosine similarity to guide cluster formation for personalized model updates; Module \ding{204} in central server: Prior Knowledge-assisted Classifier Fine-tuning (PKCF) that leverages aggregated global knowledge to enhance local classifiers' ability to discern tail classes in the server side. MuPFL synergizes these modules to capture key features, align local models, and refine classifiers—resulting in personalized local models adept at handling non-IID and long-tailed data challenges. 

\setlength{\tabcolsep}{6pt}
\begin{table}[t]
\centering
\small
\caption{\textbf{Effect of different numbers of model clustering.} Testing accuracy on CIFAR-10 with non-i.i.d degree $\alpha$=1.0 and long-tailed setting IF=10.}
\resizebox{0.5\textwidth}{!}{%
\begin{tabular}{lcccc}
    \toprule
    Method $\backslash \  \# $ of clusters& 1 &2 & 4  & 6\\
    \midrule
     FedAvg  w/o clustering& 40.41  & -
                & - & - \\
    FedAvg with clustering & -& 40.97 &  45.16 &  43.28 \\
    MuPFL       & - &72.14 &  74.55 & 73.39  \\
    \bottomrule
  \end{tabular}
}
\label{motivation table}  
\end{table}

Comprehensive experimental evaluations, including image classification and Cityscapes semantic segmentation, attest to MuPFL's efficacy in accelerating training and significantly boosting accuracy within mobile computing environments, particularly in autonomous driving scenarios, thereby outperforming state-of-the-art baselines. The contributions can be summarized as follows: 
\begin{itemize}
    \item In this work, we address the dual challenges of non-i.i.d. and long-tailed data distributions by introducing a multi-level personalized federated learning framework \texttt{MuPFL} to enhance on-device learning in resource-constrained mobile computing systems.
    \item At the local client level, we introduce the Biased Activation Value Dropout (BAVD) that selectively retains critical features according to real-time, client-specific activation maps, strategically mitigating overfitting and significantly accelerating training. 
    \item At the interface level, we propose Adaptive Cluster-based Model Update (ACMU) deployed to dynamically cluster models based on their weight similarity in each communication round, facilitating model pre-updates among clients with analogous data traits, and forging a path toward a more comprehensive global model.
    \item At the global level, we propose Prior Knowledge-assisted Classifier Fine-tuning (PKCF) to leverage collective intelligence and refine the model classifier. This strategy ensures unbiased outputs from local models, proficiently addressing the intricacies associated with non-i.i.d. and long-tailed data distributions.
\end{itemize}

\section{Related works}
\subsection{Federated learning}
The difference between FL and distributed machine learning is that data preserved on clients are non-i.i.d., and the clients' contributions are severely imbalanced, with much work being done to overcome these problems. Sattler \emph{et al.}~\cite{sattler2020clustered} introduced a clustered FL framework to generate a particular model for each cluster by using a hierarchy bi-partition algorithm. Briggs \emph{et al.}~\cite{briggs2020federated} set up experiments with different distance metrics and linkage mechanisms to divide clients into different clusters. Yao \emph{et al.}~\cite{yao2019towards} proposed FedFusion combining two model features to achieve better performance while sacrificing part of the computation power to determine the optimal fusion strategy. Wang \emph{et al.}~\cite{wang2020optimizing} proposed a mechanism based on deep Q-learning learning to determine which clients can participate in FL in each round to decrease the communication rounds needed for reaching the target accuracy. Deng \emph{et al.} ~\cite{deng2021fair} proposed a novel system to encourage clients with better local training quality to participate in federated learning and a new global aggregation algorithm. FedDyn~\cite{feddyn} adds a regularization term in local training based on the global model and the model from previous rounds of communication to overcome device heterogeneity. FedNH~\cite{dai2023tackling} employs a prototypical classifier and applies a smoothing aggregation approach for refining the classifier using clients' local prototypes.

However, such existing works leverage a step-by-step clustering strategy where each step determines a fixed cluster without flexibility during the entire training process. Our proposed ACMU is able to dynamically assign clients to different clusters.

\begin{figure*}[t]
\centering
\subfigure[Optimization path for federated learning with two clients.]{\includegraphics[width=0.525\textwidth]{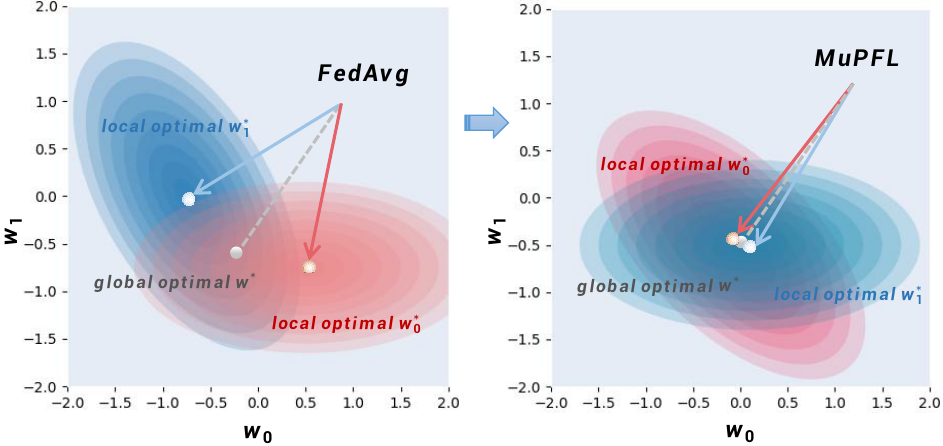}}
 \subfigure[Prediction probabilities of each class.]{\includegraphics[width=0.45\textwidth]{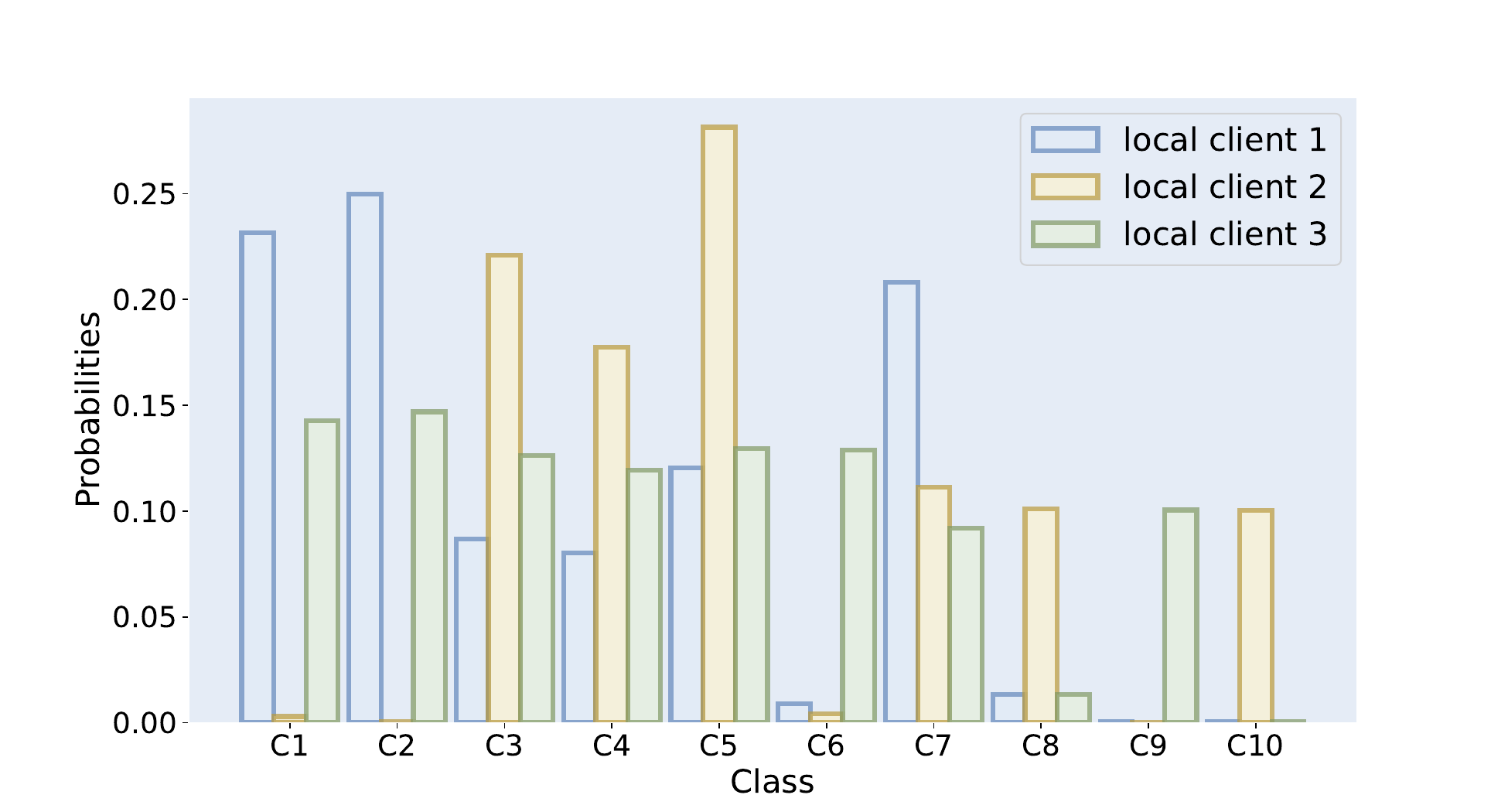}}
\caption{In sub-figure (a), it shows the optimization path for federated learning with two clients (color in cyan and red) in the case where clients' update weights diverge (left) and similar (right). For FedAvg, two clients' update weights are diverged, which intuitively indicates that forcing model aggregation can easily trap the model in a local optimum. As for MuPFL, there is a region (marked in gray in the graph) where both client's risk functions are minimized (e.g., ${w}^{*}$) thus leading to a global optimal. Sub-figure (b) is the motivation experiments conducted on CIFAR-10.}
\label{fig:motivation}
		% \vspace{-1ex}
\end{figure*}

\subsection{Long-tailed learning}
A plethora of research endeavors within the domain of deep learning has been dedicated to addressing the long-tailed distribution problem. Typically, these works can be categorized into the following distinct streams~\cite{zhang2023deep}:

\noindent\textbf{Augmentation.} Kim \emph{et al.}~\cite{kim2020m2m} used transfer learning, so that the samples of the head class are modified into samples of the tail class, and then added to the original data set to achieve the purpose of balancing the data set. Chu \emph{et al.}~\cite{chu2020feature} divided image features into two categories using CAM, one is class-related features, and the other is class-independent features and finally migrates class-independent features to the tail category for data augmentation. The FEDIC proposed by Shang \emph{et al.}~\cite{shang2022fedic} used knowledge distillation technology to transfer knowledge from the integrated model to the global model without prior knowledge of the global class distribution, and uses logit adjustment to eliminate the deviation of the integrated model.

\noindent\textbf{Decoupling.} It has been found that decoupling feature learning and classifier learning, dividing unbalanced learning into two stages, normal sampling in the feature learning stage, and balanced sampling in the classifier learning stage can bring better long-tail learning results. Kang \emph{et al.}~\cite{kang2019decoupling} initially decoupled the classification network into two parts: representation learning and classification, and systematically studied the impact of these two parts on the Long-tailed problem. It was found that using the simplest random sampling to learn representations and only adjusting the learning of the classifier can achieve a good performance.

\noindent\textbf{Re-weighting.} To address the class imbalance, re-weighting adjusts the training loss values for different classes. Lin \emph{et al.}~\cite{lin2017focal} proposed a modified focal loss based on the standard cross-entropy loss, which can reduce the weight of easy-to-classify samples so that the model can focus more on difficult-to-classify samples during training. Cao \emph{et al.}~\cite{cao2019learning} proposed Label Distribution-Aware Marginal Loss (LDAM), which defers re-weighting until after the initial stage while avoiding some of the complexities associated with re-weighting or re-sampling. Park \emph{et al.}~\cite{park2021influence} proposed an influence balance loss based on sample-aware influences to solve the overfitting of the head class in the class imbalance problem. 

\noindent\textbf{Ensemble Learning.} Ensemble learning\cite{dong2020survey,krawczyk2017ensemble,zhang2024vecaf,zhang2023unimodal} is to combine multiple weakly supervised models in order to obtain a better and more comprehensive strongly supervised model. The underlying idea of ensemble learning is that even if one weak classifier gets an incorrect prediction, other weak classifiers can correct the error. BBN~\cite{zhou2020bbn} proposed a Bilateral-Branch Network to achieve both the characterization and classification capabilities of the network, and a training strategy called cumulative learning strategy is proposed to help the network first learn the common patterns in the dataset, and then gradually focus on the tail data. Later, LTML~\cite{guo2021long} used logit statistics to correct logit, since the mean of the head logit is generally larger when uniform sampling. The bilateral-branch network scheme was also adopted to solve the long-tail problem based on BBN.

The previous work can rarely be applied to realistic non-i.i.d. and long-tailed FL scenarios. At the same time, some of these works have excessive computational overhead for local clients with inefficient training processes. In this work, we take advantage of personalized FL and incorporate novel methods to non-i.i.d. and long-tailed problems into our lightweight framework named \texttt{MuPFL} to achieve a PFL that is efficient and applicable for real-world scenarios.

\section{Methodology: MuPFL}
% Based on empirical analysis, we partition the deep learning classification model utilized in federated learning into distinct components: feature extractors and classifiers, optimizing each individually. 
% As delineated in ~\cref{framework}, our devised framework, \texttt{MuPFL}, comprises three principal modules: ~\cref{moti} elucidates the impetus behind the conception of the MuPFL framework. In ~\cref{preliminary}, we delineate the foundational notations and the optimization objectives tailored for personalized federated learning amidst non-i.i.d. and long-tailed data distributions. Detailed methodologies for Biased Activation Value Dropout, Adaptive Cluster-based Model Update and Prior Knowledge-assisted Classifier Fine-tuning are expounded in subsequent sections, referenced as ~\cref{BAVD}, ~\cref{ACMU}, and ~\cref{PKCF}, respectively. We further provide the computational complex analysis in ~\cref{analysis} and convergence analysis in ~\cref{sec:convergeana}.

\begin{figure*}[t]
        \centering
        \includegraphics[width=0.95\textwidth]{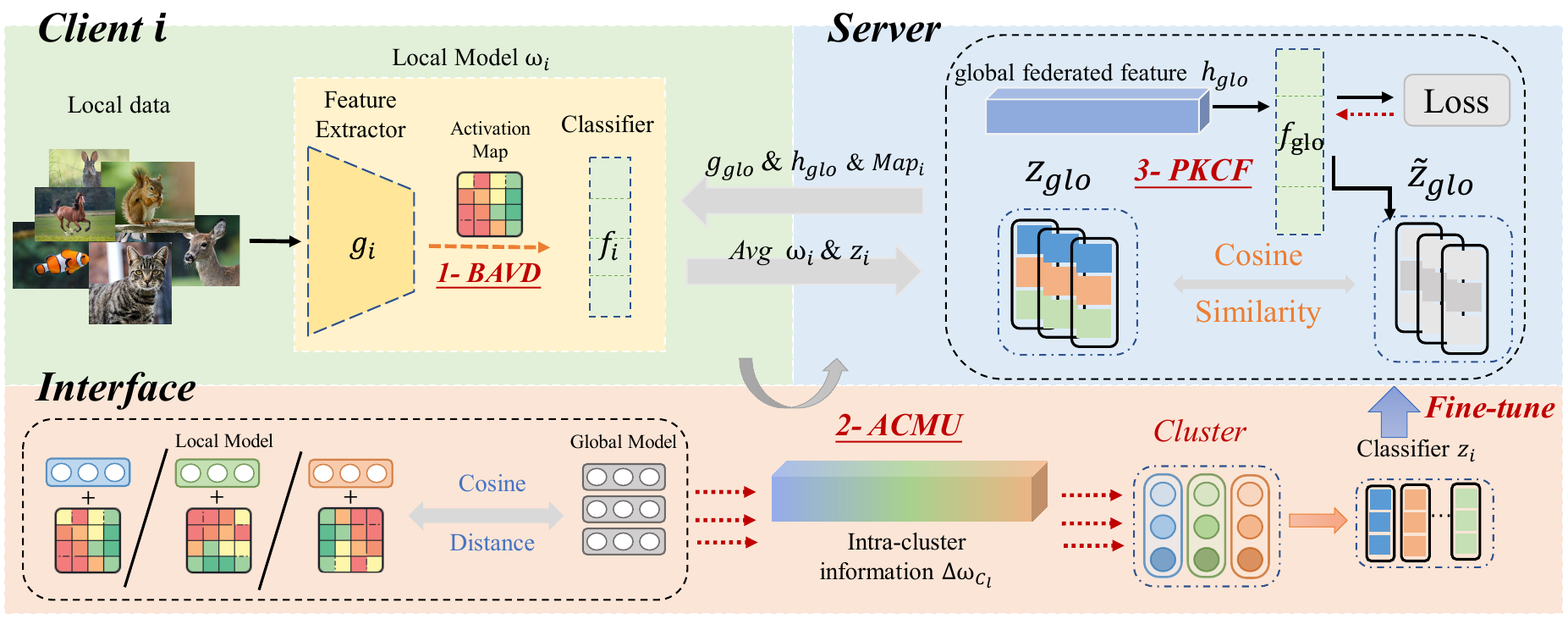}
        \caption{We propose a personalized federated learning framework \texttt{MuPFL} for non-i.i.d. and Long-tailed data. It forms a training pipeline with three stages:  1) \textit{BAVD}: local models use biased activation value dropout to tackle the overfitting problem; 2) \textit{PKCF}: clients use global federated feature $h_{glo}$ to pre-train local classifiers; 3) \textit{BAVD}: fine-tuning clients' feature extractor $g_{i}$.}
        \label{framework}
  
		% \vspace{-2ex}
\end{figure*}

In this section, we illustrate our motivation to propose MuPFL in~\cref{moti} and delineate the foundational notations and the optimization objectives tailored for personalized federated learning in non-i.i.d. and long-tailed data distributions. Moreover, we provide the in-depth description for each component in~\cref{BAVD},\cref{ACMU}, and \cref{PKCF} as delineated in ~\cref{framework}. Last but not least, we present the computational complex analysis in section~\cref{analysis} and conduct the convergence analysis in ~\cref{sec:convergeana}.

\subsection{Motivations}
\label{moti}
We aim to elucidate the impetus for our clustered knowledge-sharing strategy for personalized federated learning. The intrinsic drive for FL participation is to construct a global model that outperforms isolated local training efforts~\cite{mcmahan2017communication,2018Fedprox}. Diverging from this, evidence suggests that personalized models trained on non-i.i.d. distributions often outdo their global counterparts, a trend magnified amidst long-tailed data and pronounced client variability~\cite{fedrod,shang2022fedic}. This necessitates a nuanced, cluster-based knowledge exchange.

Our research scrutinizes the performance of a clustered federated learning approach on the long-tailed CIFAR-10 dataset~\cite{krizhevsky2009learning}, in comparison to the FedAvg baseline~\cite{mcmahan2017communication}. We concentrate on optimizing the configuration of model clusters over a network of 50 clients, as detailed in ~\cref{motivation table}. By employing cosine distances for model clustering, we observe a performance improvement ranging from 3\% to 5\%. This enhancement underscores the critical influence of the number of clusters on accuracy, while also acknowledging the added complexity this parameter introduces to the model optimization process.

Visualization of cluster-based updates are illustrated in ~\cref{fig:motivation} (a), where we delineate the optimization trajectories of the FedAvg algorithm in juxtaposition with our \texttt{MuPFL} variant under conditions of both convergent and divergent client updates. Notably, ~\cref{fig:motivation} (a-left) elucidates the shortcomings of simple averaging techniques in the context of data with a skewed distribution, often leading to convergence toward local minima. In contrast, ~\cref{fig:motivation} (a-right) exhibits the efficacy of \texttt{MuPFL}'s strategy of preemptive clustering in counteracting such distortions by guiding the algorithm towards a convergence zone (depicted as the gray area in the graph) that minimizes risk functions and steers the model toward an optimal solution.

\begin{figure}[t]
    \centering
    \includegraphics[width=0.495\textwidth]{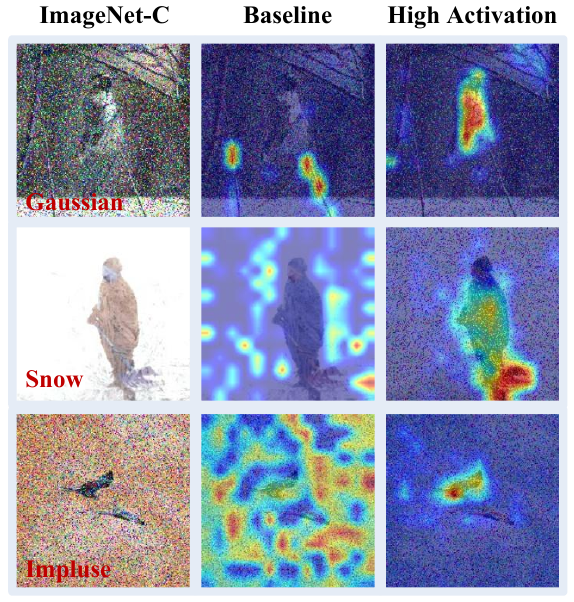}
    \caption{The visualization analysis of the Class Activation Map (CAM). We adopt CAM to compare the attention of the baseline and our proposed MuPFL with BAVD focus on the main objects in the images.}
    \label{fig:cam}
\end{figure}

We further employ the Class Attention Map (CAM) to visualize focus areas for both the baseline FedAvg and the modified FedAvg, which retains only high activation values on ImageNet-C, as shown in ~\cref{fig:cam}. Remarkably, even in environments heavily corrupted by Gaussian noise, adverse weather conditions, and impulse noise, the latter model consistently directs its attention toward the most critical regions of the image. This precise targeting significantly boosts the model's effectiveness for downstream tasks, inspiring us to explore strategic approaches for biased activation dropout.

Last but not least,the classifier outputs as visualized in ~\cref{fig:motivation} (b), reveal significant uncertainty within the model predictions, which reflects the classifiers' difficulties in handling long-tailed data distributions. Such uncertainty can be attributed to the limitations of the FedAvg algorithm, which does not adequately correct for biases induced during the training process, thereby increasing the entropy of the model's predictions. This observation substantiates the rationale behind our adoption of cluster-based updates, which are designed to mitigate bias and augment the robustness of local Deep Neural Network (DNN) classifiers. Our ultimate goal is to refine performance for practical deployments, particularly in complex real-world scenarios such as autonomous vehicular navigation.

\subsection{Preliminary}
\label{preliminary}
In the realm of federated learning, each participant denoted as client $i$, possesses a unique local dataset $D_i$, encapsulating a distribution $\mathit{\mathcal{P}_{i}\left ( \mathcal{X},\mathcal{Y} \right )}, \mathcal{Y}\in \left | \mathcal{K}\right | $, where $\mathcal{X}$ and $\mathcal{Y}$ denote the input samples and corresponding class labels. The distributions $\mathcal{P}_i$ inherently differ amongst clients, reflecting the statistical heterogeneity characteristic of real-world FL environments, a phenomenon commonly referred to as the non-i.i.d. issue. In this paper, we address scenarios wherein $\mathcal{P}_i$ manifests as a long-tailed distribution.

The classification paradigm, as embodied by model $\mathcal{F}$ with parameter set $\boldsymbol{w}$, establishes a function that maps input vectors to the label space, formalized as $y = \mathcal{F}(x; \boldsymbol{w})$, where $y$ resides within the cardinality of the label set $|\mathcal{K}|$. More specifically, the DNN bifurcates into two principle modules: 1) the \textbf{feature extractor} $f_i(x; \boldsymbol{u}_i)$, which transmutes each input instance $x$ into a latent feature representation $h \in \mathbb{R}^{B \times C \times H \times W}$; 2) the \textbf{classifier} $g_i(h; \boldsymbol{v}_i)$, which translates the extracted features into a logits vector, reflective of the confidence scores across the class spectrum. Thus, the operational mechanism of the classification model can be succinctly encapsulated as follows:
\begin{equation}
\mathbf{\hat{y}= \mathcal{F}\left ( x;\boldsymbol{w}_{i} \right ) = g_{i}\left ( h;\boldsymbol{v}_{i} \right ) \circ f_{i}\left ( x;\boldsymbol{u}_{i} \right ),}
    \label{gfpart}
    \end{equation}
where the notations $\boldsymbol{u}_i$ and $\boldsymbol{v}_i$ correspond to the parameters of the feature extractor and classifier, respectively, for the $i^{th}$ local model. Denote $\ell_i: \mathcal{Y} \times \mathcal{Y} \rightarrow \mathbb{R}$ as the cross-entropy function, pertinent to client $i$. Consequently, we express the loss function as:
    \begin{equation}
        \mathit{\mathcal{L}}_{i}(\boldsymbol{w}_{i}; D_{i})=\frac{1}{\left | D_{i} \right | }  {\sum_{j=1}^{\left | D_{i} \right|}}\ell_{i}(\mathcal{F}\left ( x_{j};\boldsymbol{w}_{i} \right );y_{j})
    \label{celoss}
    \end{equation}
In round $t$, for every participating client $i$, the stochastic gradient descent (SGD) algorithm is employed as delineated in Eq.~\ref{celoss}, aiming to minimize the loss function :
    \begin{equation}
        \boldsymbol{w}_{i}^{t+1} \gets \boldsymbol{w}_{i}^{t} -\eta \nabla_{\boldsymbol{w}} \mathcal{L}_{i}(\boldsymbol{w}_{i}; D_{i})
    \label{local}
    \end{equation}
and converge to a local optimal $\boldsymbol{w}_{i}^{*}$.Personalized federated learning encapsulates the process of jointly training a bespoke model for each client utilizing their unique dataset. Consequently, the objective within our framework is to optimize:
    \begin{equation}
    \mathop{\arg\min}\limits_{\mathcal{W}}\frac{1}{n} \sum_{i=1}^{n} \mathbb{E}_{\mathit{D_{i}}\sim \mathit{\mathcal{P}_{i}}}\left [ \ell_{i}(\mathcal{F}\left ( \mathcal{X}_{i};\boldsymbol{w}_{i} \right );\mathcal{Y}_{i} \right ] 
    \end{equation}
where $\mathcal{W}\triangleq\left \{\boldsymbol{w}_{i} \right\}_{i=1}^{N}$. Upon receiving the locally updated models from the clients, the central server conducts a global aggregation of the model parameters by:
    \begin{equation}
        \boldsymbol{w}_{glo}^{t+1} = \frac{\left|D_i\right|}{\sum_{j = 1}^{N}\left|D_j\right|} \sum_{i = 1}^{|N|} \boldsymbol{w}_i
    \end{equation}
where $N$ represents the number of local clients.

\subsection{Biased Activation Value Dropout}
\label{BAVD}
On resource-constrained clients, mitigating overfitting to boost model performance and maintain training efficiency is crucial. As dropout have often been applied to reduce overfitting in deep neural networks, its capacity to sift through and retain important information is frequently overlooked. For example, Poernomo et al.~\cite{poernomo2018biased} introduced Biased and Crossmap Dropout to categorize hidden units by size and modulate pixel correlations across feature maps with a uniform dropout pattern. In graph-based recommendation systems, Wang et al. ~\cite{wang2019neural} unveiled Message and Node Dropout to enhance representations by selectively blocking Laplacian matrix nodes with preset probabilities. However, such blanket feature omission strategies risk considerable information loss due to manual hyperparameter adjustments. To address the challenge described, we introduce an adaptive Biased Activation Value Dropout (BAVD) that selectively preserves features with the highest values based on a client-specific activation map that tracks cumulative historical feature propagation.

Within the BAVD framework, we let each client maintain an activation map which is a zero-initialized tensor $\mathcal{M}_{i}\in \mathbb{R}^{ H \times W}$, which captures the historical significance of each hidden feature $h_{i} \in \mathbb{R}^{B \times C \times H \times W}$ in each training iteration. This significance is quantified by the product of the feature's accumulated value during training and a dynamic coefficient $\psi^{(\lambda)} = \ell_{i}^{(\lambda)} - \ell_{i}^{(\lambda-1)}$, with $\lambda$ representing the specific training iteration for each batch within a given local epoch. For a discrete activation neuron $h_{ibcd} \in \mathbb{R}^{1 \times 1}$, corresponding to the $d^{\text{th}}$ element of the tensor $h_{ibc} \in \mathbb{R}^{H \times W}$ within the $c^{\text{th}}$ channel of the $b^{\text{th}}$ batch, the composition of the activation map is formulated as follows:
\begin{equation}
\mathcal{M}_{ibcd}^{(\lambda)} = \left\{
\begin{aligned}
& h^{(\lambda)}_{ibcd}, \quad \lambda=0\\
&\mathcal{M}^{(\lambda-1)}_{ibcd} + \psi^{(\lambda)} \cdot h^{(\lambda)}_{ibcd}, \quad \lambda \geq 1
\end{aligned}
\right.
\end{equation}
Normalization is applied to each activation map $\mathcal{M}_{i}\in \mathbb{R}^{H \times W}$ to preclude deviation, with its mean $\mu_{i}=\text{mean}(\mathcal{M}^{norm}_{i})$ utilized as a threshold. Activation values $h_{icbd}$ below this threshold $\mu_{i}$ are zeroed ($h_{icbd}\to 0$) within the forward pass feature map $\mathcal{M}^{norm}_{i}$. A negative coefficient $\psi^{(\lambda)}$ indicates detrimental prediction outcomes, while a positive coefficient $\psi^{(\lambda)}$ signifies contribution to model training. The activation map thus serves as a navigational guidance during feature propagation. During implementation, we substitute the standard dropout layer with our BAVD layer after the activation function in different layers.

BAVD not only accelerates the overall training speed but also significantly enhances model efficacy, offering a tailored approach over random or handcrafted selections. Additionally, the activation maps capture local data biases that lead to model biases, informing a more rational model clustering strategy within further proposed method.

% \vspace{-1ex}
\subsection{Adaptive Cluster-based Model Update}
\label{ACMU}
In our preliminary research stage in~\cref{moti}, we discover that clustering models with similar weights for updates can steer gradients towards a superior solution. Therefore, we propose Adaptive Cluster-based Model Update (ACMU) to discern local model biases by assessing the similarity between client weights and activation maps (which reveal the data distributions) and divide them into distinct clusters for preliminary updates prior to global aggregation at the FL interface stage, like edge servers. 

Therefore, the pivotal challenge becomes how to measure the similarity and how to do the clustering. We first define the updated weights as follows:
\begin{equation}
    \triangle{w^{(t)}_{i}} := w^{(t)}_{i} - w^{(t-1)}_{i}.
\end{equation}
Then we calculate the cosine similarity between two local models  $\mathit{i}$ and $\mathit{j}$ and their activation maps ($\mathit{i\neq j} $) with a hyperparameter $\alpha$ for a more comprehensive similarity evaluation: 
% 	In equation (5), we can get the update weight $\mathit{\triangle \theta_{i}^{(t)}}$ of every client $\mathit{i}$, thus we can calculate the cosine similarity between two clients  $\mathit{i}$ and  $\mathit{j}$ ($\mathit{i\neq j} $):
\begin{footnotesize}
	\begin{equation}
		\begin{aligned}
			\mathit{\mathit{Dist}^{(t)}(i,j) := \alpha cos(\triangle \boldsymbol{w}_{i}^{(t)}, \triangle \boldsymbol{w}_{j}^{(t)})} + (1-\alpha)cos(\mathcal{M}_{i}^{(t)},\mathcal{M}_{j}^{(t)})\\=\alpha \frac{\left \langle \triangle \boldsymbol{w}_{i}^{(t)},\triangle \boldsymbol{w}_{j}^{(t)} \right \rangle }{\left \| \triangle \boldsymbol{w}_{i}^{(t)} \right \|\left \| \triangle \boldsymbol{w}_{j}^{(t)} \right \|  } + (1-\alpha) \frac{\left \langle \mathcal{M}_{i}^{(t)},\mathcal{M}_{j}^{(t)} \right \rangle }{\left \| \mathcal{M}_{i}^{(t)} \right \|\left \| \mathcal{M}_{j}^{(t)} \right \|  } .\\
		\end{aligned}
	\end{equation}
 \end{footnotesize}

With cosine similarity among clients, we can further compute the Silhouette Coefficient ~\cite{rousseeuw1987silhouettes} to find the ideal cluster number $\mathit{\kappa^{(t)}}$ for client clustering. We assume that clients have been divided into $\mathit{\kappa^{(t)}}$ ($\mathit{\kappa_{min} \le \kappa^{(t)} \le \kappa_{max}}$) clusters using k-means algorithm. Moreover, for every client $\mathit{i\in C_{l}}$, we want to quantify how well client $\mathit{i}$ is assigned into cluster $\mathit{C_{l}}$. First, we denote intra-cluster dissimilarity $\mathit{a(i)}$ as:
\begin{eqnarray}
    \mathit{ a(i) = \frac{1}{\left | C_{l} \right |-1 } \sum_{j\in C_{l},i\neq j}\mathit{Dist} (i,j)^{(t)}},
\end{eqnarray}
where $\mathit{a(i)}$ is the mean distance between client $\mathit{i}$ and all other clients in the same cluster $\mathit{C_{l}}$, $\mathit{\left | C_{l} \right | }$ is the number of clients in the cluster $\mathit{i}$. Next, we define inter-cluster dissimilarity, which is the smallest mean distance of client $\mathit{i}$ to all other clients in any other cluster as:
\begin{eqnarray}
    \mathit{b(i) = \underset{k\neq l}{min} \frac{1}{\left | C_{k} \right | } \sum_{j\in C_{k}}\mathit{Dist} (i,j)^{(t)}  } .
\end{eqnarray}
Now, we can calculate a silhouette value of client $\mathit{i}$ as follows:	  
\begin{eqnarray}
    \begin{aligned}
        \mathit{S(i) = \frac{b(i)-a(i)}{max\{a(i),b(i)\}} } , if   \left | C_{i} \right | >1 \\
        \mathit{(if  \left | C_{l} \right | =1,S(i)=0)}.
    \end{aligned}
\end{eqnarray}
Then, the silhouette value of cluster number $\mathit{\kappa^{(t)}}$ is given by:
\begin{eqnarray}
    \mathit{S_{\kappa^{(t)}} } = \frac{1}{n}\sum^{n}_{i=0}S(\mathit{Dist} (i,j)^{(t)}).
\label{ideal kappa}
\end{eqnarray}
The larger the silhouette value is, the more reasonable the number of clusters to be divided, thus our goal is:
\begin{eqnarray}
    \mathit{arg max_{2\le \kappa^{(t)}\le \kappa} S_{\kappa^{(t)}}}   .
\end{eqnarray}

Through Eq.~\ref{ideal kappa}, we can derive a ideal $\mathit{\kappa^{(t)}}$ and divide clients into  $\mathit{\kappa^{(t)}}$ clusters. As illustrated in ~\cref{fig:motivation}, clients with similar updated weights $\mathit{\triangle{\boldsymbol{w}}^{(t)}_{i}}$ can converge to a more ideal solution $\boldsymbol{w}^{\ast}$.  Therefore, we will gather similar $\triangle{\boldsymbol{w}}^{(t)}_{i}$ to update the model weights of each client within the cluster. For every client $\mathit{i}$, we do cluster aggregation within each cluster as follows: $\mathit{C_{l}}$:
\begin{eqnarray}
    \mathit{\tilde{{\boldsymbol{w}}}^{(t)}_{i} = \boldsymbol{w}^{(t-1)}_{i}+  \frac{1}{\left | C_{l} \right | } \sum_{\kappa=1}^{{\left | C_{l} \right | }}\triangle{\boldsymbol{w}}^{(t)}_{\kappa}   }.  
\end{eqnarray}
By averaging the $\mathit{\triangle{\boldsymbol{w}}^{(t)}_{i}}$ inside the cluster, we can obtain a more ideal updated local model $\tilde{\boldsymbol{w}}^{(t)}_{i}$ for global aggregation. Therefore, the final global model in communication round $t$ can be obtained by:
\begin{equation}
\boldsymbol{w}^{(t)}_{global} = \sum_{i=0}^{N} \frac{n_{i}}{n} \tilde{\boldsymbol{w}}^{(t)}_{i}.
\end{equation}

Pre-sharing model parameters can mitigate the excessive model bias stemming from non-i.i.d. data issues. ACMU thus facilitates implicit information sharing among in-cluster models with closely aligned parameter distributions, promoting a more equitable and generalized global model through aggregation.

\subsection{Prior Knowledge-assisted Classifier Fine-tuning}
\label{PKCF}

We further leverage the Prior Knowledge-assisted Classifier Fine-tuning (PKCF) to strategically utilize global prior knowledge to enhance local model training in MuPFL. By aggregating heterogeneous knowledge from global users at the feature level, PKCF adeptly counteracts the imbalance caused by long-tail data distributions, a significant departure from traditional parameter-level model aggregation. 

At each round $\mathit{t}$, each client $\mathit{i}$ samples the local dataset $D_{i}$ of each class and use local feature extractor $f_{i}\left (\boldsymbol{u}_{i} \right )$  formulated in ~\cref{preliminary} to produce the real feature $h^{k} = \left \{ f_{i}\left ( x_{j};\boldsymbol{u}_{i} \right )  \right \} |_{j=1}^{\left | D_{i} \right | }, h^{k} \in \mathbb{R}^{B \times C \times H \times W}$ for each class $k$. Thus, the real feature gradient $z_{i}^{k}$ of each class $k$ can be computed as follow:
\begin{equation}
    z_i^k=\frac{1}{\left|D_i^k\right|} \sum_{j=1}^{\left|D_i^k\right|} \nabla_{\boldsymbol{v}_i} \ell\left(g_i\left(h_i ; \boldsymbol{v}_i\right), y_j\right),
\end{equation}
and we denote $z_i^{t+1}=\left\{z_i^k \mid k \in \left | \kappa \right |\right\}$. Due to the imbalance in the number of each class, we have $ \kappa  \subseteq\left | \mathcal{K} \right |$. Each selected client $\mathit{i}$ then transfers $z_i^{t+1}$ to the server. Such transmission of feature gradients brings benefits in terms of privacy preservation since they are averaged by each class, which is an irreversible process. 

\begin{algorithm*}[t]  
  \caption{The framework of \texttt{MuPFL}}  
  \begin{algorithmic}[1] 
    \Procedure {MuPFL}{} \Comment{On server}
    
        \State {$\mathbf{Initialization}: \boldsymbol{w}_{0}, \mathit{D}, \mathit{N}, \mathcal{M}_{0}$\;}
        \For {\textit{communication round} $t \in [1,T]$}
        \State {$S_{t} \gets$ random select subset of m clients} \Comment{Select a set of participated clients}

            \For {$client i \in S_{t}$}
            % \State {$w^{(t)}_{i}, \mathcal{M}_{i} \gets  CLIENTUPDATE(k, BAVD(\delta, \mathcal{M}_{0}), w^{(t-1)}_{i}) \;$}
            \State {$w^{(t)}_{i} \gets  CLIENTUPDATE(k, \mathit{D}_{i}, \mathcal{M}_{i_{0}}, \boldsymbol{w}^{(t-1)}_{i}) \;$}  \Comment{Local selected clients training with BAVD algorithm} 
                    
            \State {$\triangle{\boldsymbol{w}^{(t)}_{i}} \gets \boldsymbol{w}^{(t)}_{i} - \boldsymbol{w}^{(t-1)}_{i}$}  \Comment{Local selected client compute updated weights} 
                    
        	\EndFor
                    
         \State {$\{\tilde{\boldsymbol{w}}^{(t)}_{i}\}_{i \in S_{t}} \gets \textbf{ACMU}(\{\triangle{\boldsymbol{w}}^{(t)}_{I}, \boldsymbol{w}^{(t)}_{i}, \mathcal{M}_{i}^{(t)}\}_{i \in S_{t}}) \;$}  \Comment{Local clients update within clusters} 

        \State {$\boldsymbol{w}^{(t)}_{global} \gets \sum_{i=0}^{N} \frac{n_{i}}{n} \tilde{\boldsymbol{w}}^{(t)}_{i}$}  \Comment{Global model update using Average algorithm} 
        \State{\textbf{PKCF}({$z_{i}$})}
        % \frac{\sum^{\mathit{n}}_{i=1}\mathcal{Z}_{i}^{(t)}\tilde{{w}}^{(t)}_{i}}{\sum^{\mathit{n}}_{i=1}\mathcal{Z}_{i}^{(t)}}$}
                    
        \EndFor
        \EndProcedure
        \State
        
        \Procedure {CLIENTUPDATE}{$k$, $\delta_{i}$, $\mathcal{M}_{i_{0}}$, ${w}^{(t-1)}_{i}$} \Comment{On client $i$}
        
			\State {$\mathcal{B}_{i}  \gets$ \textit{(Split} $\mathit{D}_{i}$ \textit{into batches of size }$B$)}  \Comment{Split local dataset in mini-batch for training} 
                \State{$C_i \leftarrow $\textit{ Sample subset of }  $\mathit{D}_{i}$ } \Comment{Sample a subset of local dataset for compute feature gradients} 
			
			\For {local epoch $e$ from 1 to $E$}
			
    			\For {batch $b \in \mathcal{B}$}
    			    \State{$\mathcal{M}_{i} \gets \textbf{BAVD}(CNN(b_{i}))$} \Comment{Local clients compute the activation map} 
                    \State{$\boldsymbol{w}^{(t)}_{i} \gets \boldsymbol{w}^{(t-1)}_{i} - \varphi\nabla \ell(\boldsymbol{w}^{(t-1)}_{i})$}  \Comment{Local clients update using SGD algorithm} 
                \EndFor
                \EndFor
                \For {each class sample in $C_i$ }
                \State{       $ z_i^k=\frac{1}{\left|D_i^k\right|} \sum_{j=1}^{\left|D_i^k\right|} \nabla_{v_i} \ell\left(g_i\left(h_i ; v_i\right), y_j\right)$} \Comment{Compute the real feature gradient $z_{i}^{k}$ of each class $k$}
                \EndFor
			\State \Return {$\boldsymbol{w}^{(t)}_{i}$, $\mathcal{M}_{i}$, $z_i^k$ to interface}
		\EndProcedure
        \State
		
        \Procedure {\textbf{ACMU}}{$\{\triangle{\boldsymbol{w}}^{(t)}_{I}, {w}^{(t)}_{i}\}_{i \in S_{t}}$} \Comment{On interface}
        
			\State {$\delta \gets$ cosine similarity between two different client i, j} \Comment{Calculate the cosine similarity between the pair clients}
			
			\For {$\kappa \in  [ \kappa_{min}, \kappa_{max}]$}
    			\State {$cluster_{\kappa}\gets$ use $\delta$  and divide clients into $k$ cluster} \Comment{Divide clients into cluster using cosine similarity}
    			
    			\State {{$S_{\kappa^{(t)}}$}  $\gets$  calculate silhouette value}
    			
            \EndFor
            \State {$\kappa \gets$ $\kappa$ clusters with maximum $S_{\kappa^{(t)}}$} \Comment{Choose the optimized cluster}
            
            \State {$\mathit{\tilde{\boldsymbol{w}}^{(t)}_{i} = \boldsymbol{w}^{(t-1)}_{i}+  \frac{1}{\left | C_{l} \right | } \sum_{\kappa=1}^{{\left | C_{l} \right | }}\triangle{\boldsymbol{w}}^{(t)}_{\kappa}   }$}  
            
			\State {\Return {$\{\tilde{\boldsymbol{w}}^{(t)}_{i}\}_{i \in S_{t}} $}}
			
		\EndProcedure
    \State{}
    \Procedure {\textbf{PKCF}}{$z_{i}$} \Comment{On server}
    \State{$        z_{glo}^k=\frac{1}{\left|\mathcal{S}^k\right|} \sum_{i=1}^{|\mathcal{S}|} z_i^k$}  \Comment{Aggregate feature gradients of the same class}
    \State{$        \hat{z}_{glo}^{k}=\frac{1}{m} \sum_{j=1}^m \nabla_{v_{glo}} \ell\left(g_{glo}\left(h_{glo},v_{glo}) ; y_j\right)\right.$}  \Comment{Compute the global feature gradient $\hat{z}_{glo}$ for each class $k$}
    \State{$\arg\min \cos \left(z_{glo}, \hat{z}_{glo}\right)=\frac{1}{|\mathcal{K} |} \sum_{j=1}^{\mathcal{K} }\frac{\left\langle z_{glo}^k, \hat{z}_{glo}^k\right\rangle}{\left\|z_{glo}^k\right\|\left\|\hat{z}_{glo}^k\right\|}$} \Comment{ Minimize the cosine distance between $z_{glo}^k$ and $\hat{z}_{glo}^{k}$ to obtain $h_{glo}^{t+1}$}
    \State {\Return {$h_{glo}^{t+1}$ to selected clients}} \Comment{On client} Each client using $h_{glo}^{t+1}$ to pre-train local classifier
    
    \EndProcedure

\end{algorithmic}
\end{algorithm*}  

When received $z_i^{t+1}$, the server starts to generate the global federated feature for the succeeding local classifier tuning. We first aggregate feature gradients of the same class from different client $\mathit{i}$:
\begin{equation}
    z_{glo}^k=\frac{1}{\left|\mathcal{S}^k\right|} \sum_{i=1}^{|\mathcal{S}|} z_i^k.
\end{equation}
    
Since the classification model $\mathcal{F}$ is concatenate by two modules (Eq.~\ref{gfpart}), then we have:
\begin{equation}
    \mathcal{F}_{glo}\left ( w_{glo} \right ) = g_{glo}\left ( \boldsymbol{v}_{glo} \right ) \circ f_{glo}\left (\boldsymbol{u}_{glo} \right ) 
\end{equation}
Therefore, we can initiate the global federated feature $h_{glo} = f_{glo}(\boldsymbol{u}_{glo})$ and compute the global feature gradient $\hat{z}_{glo}$ for each class $k$ iteratively:
\begin{equation}
    \hat{z}_{glo}^{k}=\frac{1}{m} \sum_{j=1}^m \nabla_{\boldsymbol{v}_{glo}} \ell\left(g_{glo}\left(h_{glo},\boldsymbol{v}_{glo}) ; y_j\right)\right. ,
\end{equation}
where $m$ is the number of features. The optimization goal of feature generation on the server is to minimize the cosine distance between $z_{glo}^k$ and $\hat{z}_{glo}^{k}$ for every class: 
\begin{equation}
    \arg\min \cos \left(z_{glo}, \hat{z}_{glo}\right)=\frac{1}{|\mathcal{K} |} \sum_{j=1}^{\mathcal{K} }\frac{\left\langle z_{glo}^k, \hat{z}_{glo}^k\right\rangle}{\left\|z_{glo}^k\right\|\left\|\hat{z}_{glo}^k\right\|},
\end{equation}
and the dimension of $h_{glo}^{t+1}$ is $|\mathcal{K} |\times m$. Later, the server distributes $h_{glo}^{t+1}$ to selected clients for local classifier tuning in the next round $t+1$. Since the $h_{glo}^{t+1}$ generated by the server has global prior knowledge, using $h_{glo}^{t+1}$ to pre-train the client's local classifier can enhance the local classifier's ability to classify small-scale samples. The new classifier of client $i$ is trained after $\tau$ epochs as follows:
\begin{equation}
    \boldsymbol{u}_i^{t+1} \leftarrow \boldsymbol{u}_i^t-\eta \nabla_{\boldsymbol{u}_i} \ell\left(g_i\left(h_{glo}^{j}; \boldsymbol{u}_i^t\right) ; y_j\right).
\end{equation}
A further discussion of the hyperparameter $m$ and $\tau$ will be presented in Experiments. After classifier tuning, client $i$ will use the new parameters $\boldsymbol{w}_{i}^{t+1}=(\boldsymbol{u}_{i}^{t+1},\boldsymbol{v}_{glo}^{t+1})$ for local model training and feature gradients computation.

\subsection{Computational complexity analysis}
\label{analysis}
In federated learning systems, computational tasks are distributed between local mobile devices and a central server. The BAVD module utilizes torch.topk for processing feature embeddings $h_{i}\in\mathbb{R}^{B\times C\times H\times W}$,  with its time complexity for spatial dimensions driven by the algorithm's sorting efficiency. Specifically, topK algorithm conducts a partial sort to locate the top $K$ values in each $H\times W$ area of the batch and channel dimensions, resulting in a complexity of $\mathcal{O}(H\times W log(K))$ per slice. Consequently, across all $B\times C$ slices, the cumulative time complexity of BAVD stands at $\mathcal{O}(B\times C\times H\times Wlog(K))$., and the computational effort required to determine the mean can be considered negligible. This is relatively minor when juxtaposed with the typical computational demands of forward and backward propagation in neural network training.

Given that both ACMU and PKCF primarily rely on cosine computations, we analyze them in conjunction. For local model weights $\boldsymbol{w}_{a},\boldsymbol{w}_{b}\in\mathbb{R}^{D\times H\times W}$, computing a single cosine distance has a complexity of $\mathcal{O}(D\times H\times W)$. The complexity for all $\frac{N(N-1)}{2}$ weight pairs is $\mathcal{O}(\frac{N(N-1)}{2}\times D\times H\times W)$, given the quadratic nature of pairwise comparisons. For silhouette assessment with $Q$ clusters among $N$ models, the per-model complexity for intra- and inter-cluster distances, $a(i)$ and $b(i)$ is $\mathcal{O}(N\times Q\times D\times H\times W)$. Hence, silhouette computation for all models is $\mathcal{O}(N^{2}\times Q\times D\times H\times W)$. The joint complexity for pairwise cosine distances and silhouette scores is then equal to $\mathcal{O}(N^{2}\times D\times H\times W)+\mathcal{O}(N^{2}\times Q\times D\times H\times W)$. Given $Q\ll N$, the overarching complexity is dominated by $\mathcal{O}(N^{2}\times D\times H\times W)$.

Consequently, the aggregate computational complexity of MuPFL, incorporating the aforementioned time complexities, is rendered as $\mathcal{O}(B\times C\times H\times Wlog(K))+\mathcal{O}(N^{2}\times D\times H\times W)$. Utilizing the formidable computational prowess of the central server, the MuPFL framework adeptly manages the resource-heavy ACMU and PKCF modules, while offloading the BAVD module to less-capable terminal mobile units. This judicious distribution of computational labor ensures that MuPFL, with its efficient architecture, is suitably equipped for deployment in autonomous driving systems.

\subsection{Convergence analysis}
\label{sec:convergeana}
In this section, we analyze the convergence bound of
MuPFL in a view of federated learning framework~\cite{li2019convergence,2018Fedprox}.

\begin{assumption}[Lipschitz Smooth and Continuity]
Each local optimization function exhibits $L_1$-Lipschitz smoothness, indicating that the gradient of each objective function maintains $L_1$-Lipschitz continuous.
\begin{equation}
\Vert\nabla \ell( \boldsymbol{w}^{(t_1)})- \nabla \ell (\boldsymbol{w}^{(t_2)} )\Vert_2 \leq L_1 \Vert  \boldsymbol{w}^{(t_1)}- \boldsymbol{w}^{(t_2) } \Vert_2,\forall t_1, t_2 > 0.
\end{equation}
This equation also implies the quadratic bound:
\begin{align}
 \ell( \boldsymbol{w}^{(t_1)})&- \ell( \boldsymbol{w}^{(t_2)}) \leq \langle \ell( \boldsymbol{w}^{(t_2)}), \boldsymbol{w}^{(t_1)}- \boldsymbol{w}^{(t_2)}\rangle 
 \\ &+ \frac{L_1}{2}\Vert  \boldsymbol{w}^{(t_1)}- \boldsymbol{w}^{(t_2)} \Vert_2^2,\forall t_1, t_2 > 0 .\notag   
 \end{align}
% In addition, each client's local feature extractor $f_i(\phi_i^{t})$ also follows $L_2$-Lipschitz continuous:
% \begin{equation}
% \Vert f_i(\phi_i^{t_1})- f_i(\phi_i^{t_2})(\cdot) \Vert_2 \leq L_2 \Vert  \phi_i^{t_1}- \phi_i^{t_2}\Vert_2,\forall t_1, t_2 > 0.
% \end{equation}
\end{assumption}

\begin{assumption}[Unbiased Gradient and Bounded Variance]
Let $\textbf{g}_i^{(t)}=\nabla \ell( \boldsymbol{w}^{(t)}_i, \xi_i^{(t)})$ be the stochastic gradient, which denotes as an unbiased estimator of the local gradient for each client. Assuming its expectation value as :
\begin{equation}
\mathbb{E}_{\xi_i^{(t)} \sim D_i} [\textbf{g}_i^{(t)}]=\nabla \ell( \boldsymbol{w}^{(t)}_i), \forall i \in\{ 0,1,\cdots, N\}.
\end{equation}
where the variance of Gradient is bounded by $\sigma^2$:
\begin{equation}
\mathbb{E}[\Vert \textbf{g}_i^{(t)}-  \ell( \boldsymbol{w}^{(t)}_i)  \Vert_2^2] \leq \sigma^2, \forall i \in \{0,1,\cdots, N\} , \ \sigma >0.
\end{equation}
\end{assumption}

\begin{assumption}[Bounded Expectation of Euclidean norm of Stochastic Gradients]
The expectation of the stochastic gradient is
bounded by G:
\begin{equation}
\mathbb{E}[\Vert \textbf{g}_i^t \Vert_2^2]\leq G, \  \forall i \in \{0,1,\cdots, N\}.    
\end{equation}
\end{assumption}

\begin{theorem}
Combining the above assumption, we can derive the convergence bound of the proposed MuPFL for a client $i$ after local training epochs $E$ at global communication round $t+1$ as:
\begin{equation}
\mathbb{E}[\ell_E^{t+1}]\leq \ell_{\frac{1}{2}}^{t+1 }-(\eta-\frac{L_1 \eta^2}{2})\sum_{e=\frac{1}{2}}^{E-1}\Vert \nabla \ell_e^{t+1}  \Vert_2^2 +\frac{\eta^2L_1E\sigma^2}{2}.
\end{equation}
\end{theorem}
The above inequality holds the theorem that the loss function converges.

\section{Experiments}

In this section, we investigate the efficacy of \texttt{MuPFL} in addressing the long-tail distribution challenge inherent in mobile computing by conducting comprehensive evaluations on established image classification and semantic segmentation benchmarks. Targeted ablation studies on the BAVD, ACMU, and PKCF modules quantify their respective impacts. The experimental results demonstrate that \texttt{MuPFL} notably outperforms contemporary state-of-the-art methods by a large extent.

\setlength{\tabcolsep}{6pt}
\begin{table*}[t]
\footnotesize
\begin{center}      
\caption{Top-1 average client test accuracy (\%) with other state-of-the-art federated learning baselines and \texttt{MuPFL} on three datasets including Fashion-MINST, CIFAR-10 and CIFAR-100 under non-i.i.d. and Long-tailed setting}
	\resizebox{2\columnwidth}{!}{%
 \begin{tabular}{ccccccccccccc}
\toprule 
\multicolumn{1}{c}{\multirow{3}{*}{Method}} & \multicolumn{4}{c}{Fashion-MNIST}  & \multicolumn{4}{c}{CIFAR-10 }     & \multicolumn{4}{c}{CIFAR-100 }
\\ 
\cmidrule(r){2-5}\cmidrule(r){6-9}
\cmidrule(r){10-13}
  & \multicolumn{2}{c}{IF=10} &  \multicolumn{2}{c}{IF=100}& \multicolumn{2}{c}{IF=10} & \multicolumn{2}{c}{IF=100}&  \multicolumn{2}{c}{IF=10}& \multicolumn{2}{c}{IF=100}\\
\cmidrule(r){2-3}
\cmidrule(r){4-5}
\cmidrule(r){6-7}
\cmidrule(r){8-9}
\cmidrule(r){10-11}
\cmidrule(r){12-13}

  & $\alpha$=0.5 & $\alpha$=1.0 & $\alpha$=0.5 & $\alpha$=1.0& $\alpha$=0.5 & $\alpha$=1.0& $\alpha$=0.5 & $\alpha$=1.0& $\alpha$=0.5 & $\alpha$=1.0& $\alpha$=0.5 & $\alpha$=1.0  \\ 
 \midrule
\textsc{FedAvg} & 78.26 & 83.34 & 81.76 & 84.22 & 40.41 & 45.63 & 52.19 & 55.97 & 20.39& 24.01 & 26.09 & 26.84   \\
\textsc{FedProx} & 86.92 & 89.57 & 90.72 & 91.56 & 58.29 & 63.43 & 67.33 & 71.88 & 31.98& 33.47 & 35.97 & 38.19   \\
\textsc{FedNova} & 86.30 & 88.03 & 88.62 & 90.21 & 45.56 & 53.76 & 56.96 & 67.26 & 23.28& 24.86 & 25.48 & 26.91   \\
\midrule
\textsc{Ditto} & 89.45 & 90.75 & 91.06 & 92.45 & 63.22 & 67.10 & 69.89& 70.16 & 33.59 & 36.85 & 37.67& 39.19   \\
\textsc{FedRep} & 89.78 & 91.09 & 91.55 & 92.77 & 64.09 & 67.86 & 70.01 & 70.48  & 34.47 & 37.01 & 38.78 & 39.86   \\
\midrule
\textsc{CReFF} & 89.98 & 91.57 & 91.73 & 92.98 & 63.51 & 67.37 & 70.33 & 73.71 & 37.43 & 39.33 & 42.53 & 45.41   \\
% \midrule
\textsc{MuPFL} &  
\textbf{91.90} & \textbf{93.89} &\textbf{94.11}& \textbf{94.39} & \textbf{70.02}& \textbf{74.76} & \textbf{75.88}& \textbf{77.09}  & \textbf{40.55}& \textbf{43.91} & \textbf{48.97}& \textbf{51.26} \\
\bottomrule
\end{tabular}
}
\label{table:acc_all}
\end{center}
\end{table*}

\begin{figure*}[t]
\centering
\subfigure[Example of imbalance data partition on CIFAR10 dataset]{\includegraphics[width=0.24\textwidth]{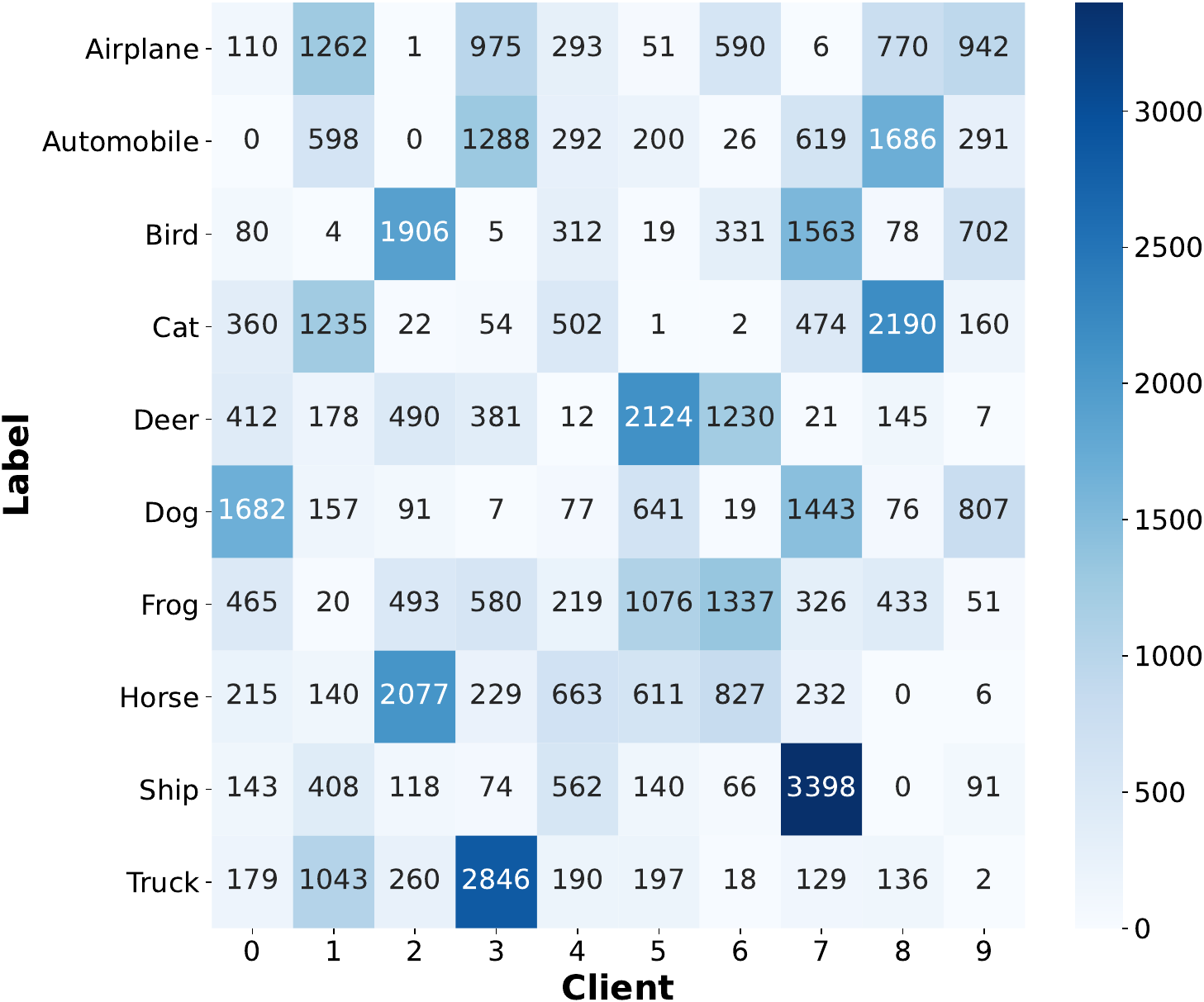}}
 \subfigure[Fashion-MNIST: Epochs \textit{v.s.} Test Accuracy]{\includegraphics[width=0.24\textwidth]{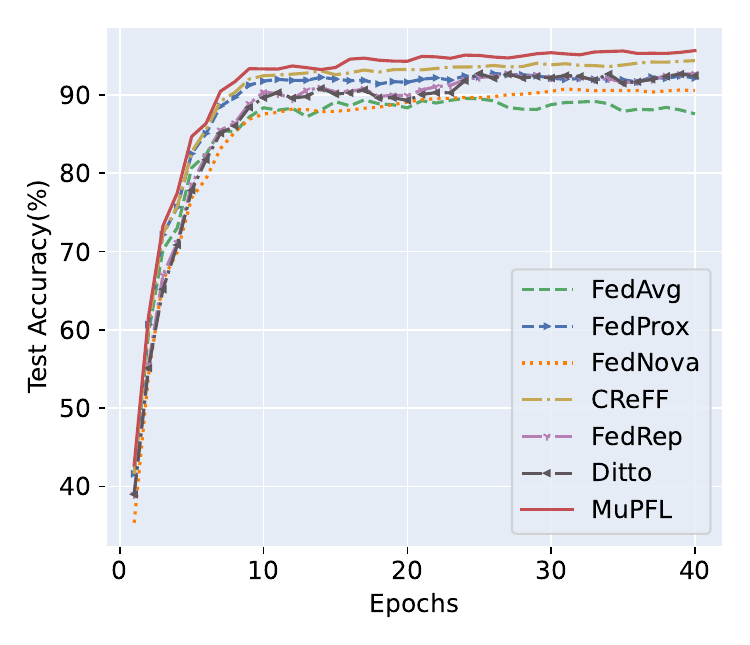}}
\subfigure[CIFAR-10: Epochs \textit{v.s.} Test Accuracy]{\includegraphics[width=0.24\textwidth]{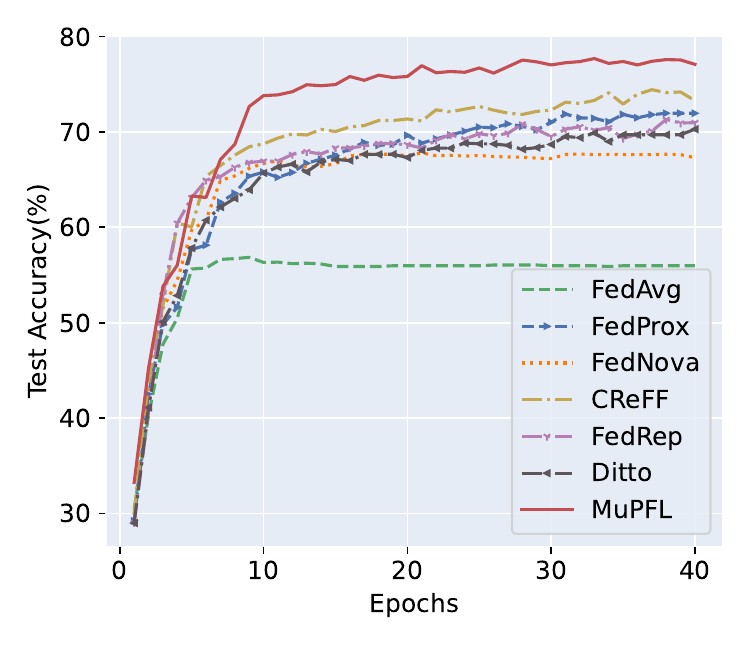}}
\subfigure[CIFAR-100: Epochs \textit{v.s.} Test Accuracy]{\includegraphics[width=0.24\textwidth]{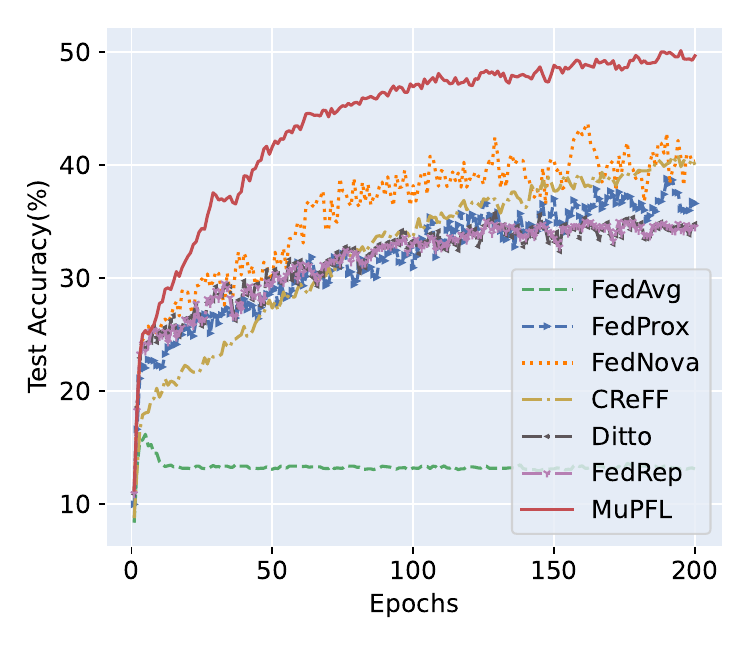}}
\caption{(a) An example of a distribution-based label imbalance partition the CIFAR10 dataset ($\alpha$,IF)=(1.0,10) with 10 clients. (b-d) The performance of different baselines over the experiments setting ($\alpha$,IF)=(1.0,10) on three datasets.}
\label{acc}
\end{figure*}

\subsection{Experimental setup}
\noindent\textbf{Datasets and models.} We conducted our experiments on a high-performance computing platform equipped with an Intel(R) Core(TM) i9-10900X CPU, utilizing 4 NVIDIA GeForce RTX 3090 GPUs, and supported by 128GB of RAM. The experimental framework was rigorously applied to two principal tasks in mobile computing:
\begin{itemize}
    \item \textbf{Image classification.} We employed three widely recognized datasets and two DNN models. It is imperative to highlight that, in alignment with mobile computing contexts, we have opted for relatively compact models. Specifically, for the FashionMNIST dataset~\cite{xiao2017fashion}, we utilize a CNN composed of two convolutional layers with 5$\times$5 kernels followed by a single linear layer. For the CIFAR-10 and CIFAR-100 datasets~\cite{krizhevsky2009learning}, we implement a streamlined version of the ResNet architecture, ResNet-8~\cite{he2016deep}, chosen for its balance between model complexity and performance suited to the computational constraints of mobile platforms.

    \item \textbf{Semantic segmentation.} utilize the Cityscapes dataset~\cite{cordts2016cityscapes}, an extensive collection of urban street scene imagery designed for comprehensive semantic analysis. This dataset is equipped with semantic, instance-level, and pixel-precise annotations across 30 categories, incorporating 5,000 images with fine annotations and an additional 20,000 images with coarse annotations. For the task of semantic segmentation, we employ the MobileNet-V3-small~\cite{howard2019searching} due to its optimal balance of efficiency and accuracy, making it particularly suitable for real-time applications in computer vision such as those required in autonomous driving systems.

\end{itemize}

\noindent As for data partition, we use Dirichlet distribution $\mathit{\operatorname {Dir} ({\boldsymbol {\alpha }})}$ to generate a different distribution of each class corresponding to each client $\alpha=\{0.5, 1.0\}$. Second, we follow the setting in ~\cite{cao2019learning} to simulate long-tailed distributions where IF indicates the ratio of major and minor categories and set IF=[10,100]. \textbf{Specifically for segmentation, Cityscapes is a semantic segmentation dataset of street view with 19 semantic classes, respectively. Unlike classification, an image from semantic segmentation datasets contains objects of many classes that are hard to split. To generate the class-heterogeneous data partition among clients, we split Cityscapes into 19 subsets. Each subset maintains $\beta$ semantic classes and sets other classes as background. In this setting, a problem arises due to the inconsistent foreground-background settings across different clients, which display various degrees of non-i.i.d. data.}

\noindent\textbf{Baselines.} To rigorously evaluate the efficacy of our proposed method, \texttt{MuPFL}, we have selected a diverse array of baseline models that specifically address challenges associated with non-i.i.d. data distributions and penalization concerns in federated learning environments.
\begin{itemize}

    \item \textbf{FedAvg}~\cite{mcmahan2017communication} is the most famous and maybe most widely-used FL system in the world. FedAvg averages the uploaded clients' model parameters and then randomly chooses the participating clients in the next communication round.

    \item \textbf{FedProx}~\cite{li2020federated} can be viewed as a generalization and re-parametrization of FedAvg, with an additional penalty loss item to handle the data non-i.i.d. scenarios.

    \item \textbf{FedNova}~\cite{NEURIPS2020_564127c0} provides a normalized averaging method to deal with data non-i.i.d. and eliminates objective inconsistency caused by na\"ive weight parameter aggregation.

    \item \textbf{Ditto}~\cite{li2021ditto} can be viewed as a lightweight personalization add-on for standard global FL to achieve personalized FL.

    \item \textbf{FedRep}~\cite{pmlr-v139-collins21a} learns a shared data representation across clients and unique local heads for each client for FL personalization.

    \item \textbf{CReFF}~\cite{shang2022federated} only re-trains the classifier on federated features and can achieve comparable model performance as the one re-trained on real data under a long-tailed data scenario.
\end{itemize}
To substantiate the robustness of \texttt{MuPFL}, we have executed a series of detailed ablation studies. These evaluations have been designed to dissect and examine the individual contributions of \texttt{MuPFL}'s components, elucidating the method's strengths from multiple dimensions.

\noindent\textbf{Implementation details.} For other algorithmic parameters, we set as follows: local and global learning rate $\eta =1e^{-2}$ and
learning rate decay rate of 0.005 every 3 epochs., local batch size equal to 64, local epoch equal to 10, global epoch equal to 40 for FashionMNIST and CIFAR-10, 200 for CIFAR-100, and 10000 for Cityscapes. We design a total of 50 (for FashionMNIST and CIFAR-10) and 19 (for CIFAR-100 and Cityscapes) clients and the fraction selected for each epoch is 20\%. As for the hyper-parameters 1) the number of feature $m$, we set to [10,50,100,150,200], 2) Fine-tuning classifier epochs, we set to [10,50,100]. All the other setting follows the same as ~\cite{howard2019searching} for MobileNet-V3-small. It should be noted that our proposed PKCF can also be applied to the segmentation head in MobileNet-V3-small.

\begin{figure}[t]
\centering
% \subfigure[Imbalance data partition.]{\includegraphics[width=0.3\textwidth]{images/test.pdf}}
\subfigure[The performance of hyper-parameter $m$]{\includegraphics[width=0.24\textwidth]{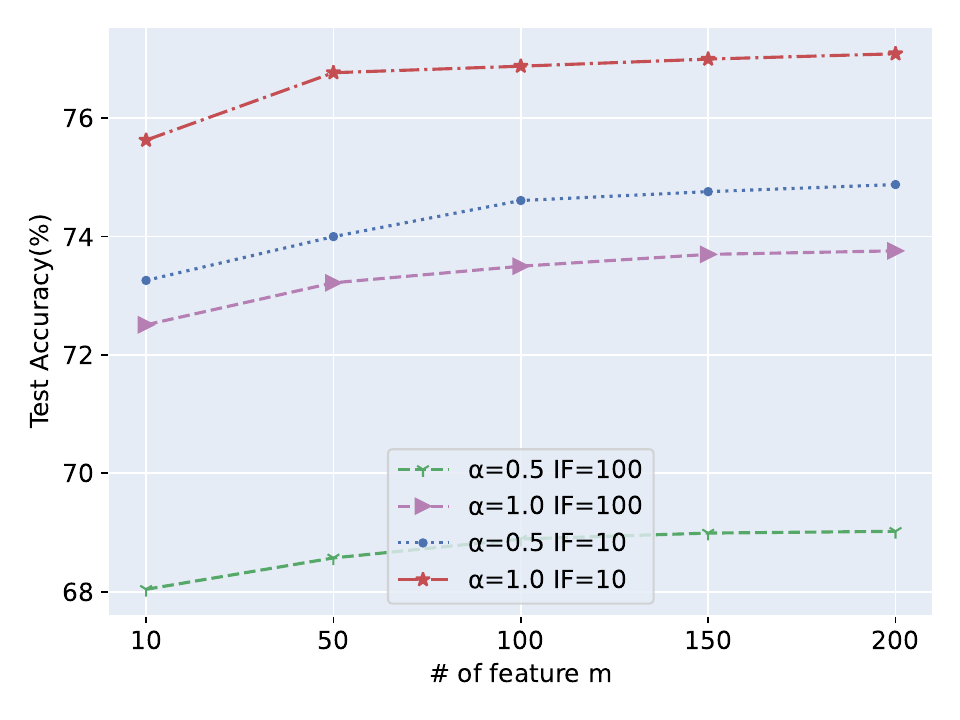}}
\subfigure[The performance of hyper-parameter $\tau$]{\includegraphics[width=0.24\textwidth]{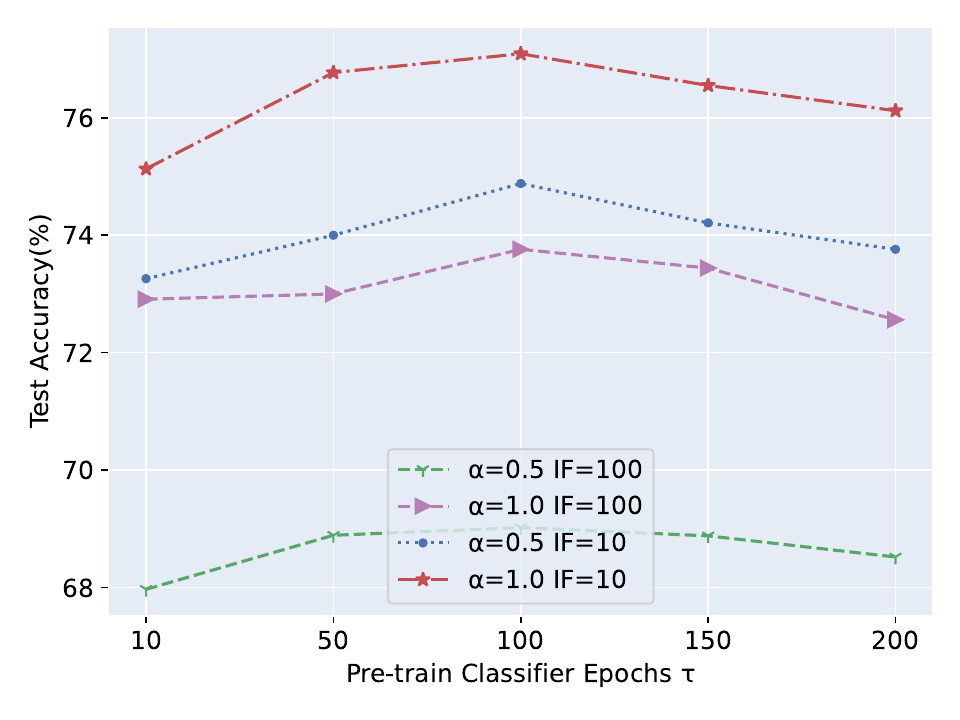}}
\caption{Influence of hyper-parameter on CIFAR-10 with different non-i.i.d scenario $\alpha$ and IF.}
\label{fig:hyper}
\end{figure}

\subsection{Quantitative analysis}
\textbf{Image Classification.} In all our experiments, we report the average testing accuracy of overall clients, as shown in ~\cref{table:acc_all}. To show the convergence rate, we illustrate how MuPFL and several baselines behave under the data partition $(\alpha$, IF)=(1.0, 10), as shown in ~\cref{acc}. We intuitively see that on the three different datasets, MuPFL has a large degree of accuracy improvement compared to all baselines.

Baselines such as FedProx and FedNova, tailored for non-i.i.d. settings, falter in long-tailed distributions, incurring substantial accuracy declines. In contrast, MuPFL not only resists this drop but also secures an accuracy boost between 2.83\% and 24.46\%. Personalized federated learning approaches like Ditto and FedRep, while generally effective, still struggle with tail class recognition due to sample scarcity. Incorporating PKCF, MuPFL harnesses global knowledge for pre-training local classifiers, enhancing tail class accuracy. Additionally, ablation studies (refer to \cref{abla}) demonstrate that both PKCF and BAVD improve the local model's fit to its dataset. MuPFL excels by facilitating higher accuracy through personalized training and mitigates local tail class paucity with global knowledge infusion.

\begin{figure*}[t]
\includegraphics[width=\textwidth]{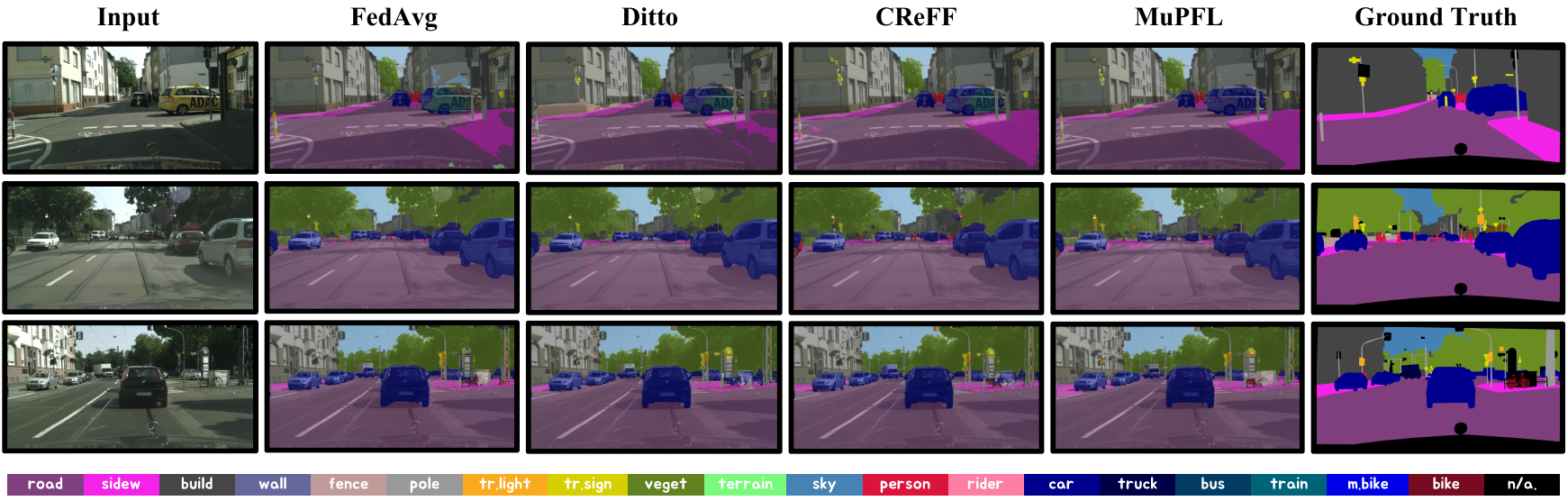}
\centering
% \vspace{-0.5cm}
\caption{Example of image input(left) ground truth(right), and qualitative results on Cityscapes with base model MobileNet-V3-small on different non-i.i.d. and long-tailed targeted federated learning baselines.}
\label{fig:vis}
\end{figure*}

\noindent\textbf{Semantic Segmentation.} 
Semantic segmentation is one of the most common tasks in autonomous driving scenarios. To further demonstrate the superiority of our proposed MuPFL in dealing with real-time open-world applications, we provide the qualitative comparison on Cityscapes for semantic segmentation tasks based on mIoU(\%) and mAcc(\%), which is shown in ~\cref{real_results}. We can find that other federated learning baselines fail to make satisfying predictions with only 44.46\% mIoU and 87.42\% mAcc for FedAvg. On the other hand, our proposed MuPFL that focuses on tackling the non-i.i.d. and long-tailed challenges provides satisfactory performance on this task by outperforming 2.58\% mIoU and 1.24\% mAcc compared with other long-tailed targeted federated learning methods. We further provide the visualization results in \cref{fig:vis}.

\begin{figure}[t]
\centering
\subfigure[Time cost to reach 70\% target accuracy on CIFAR-10]{\includegraphics[width=0.24\textwidth]{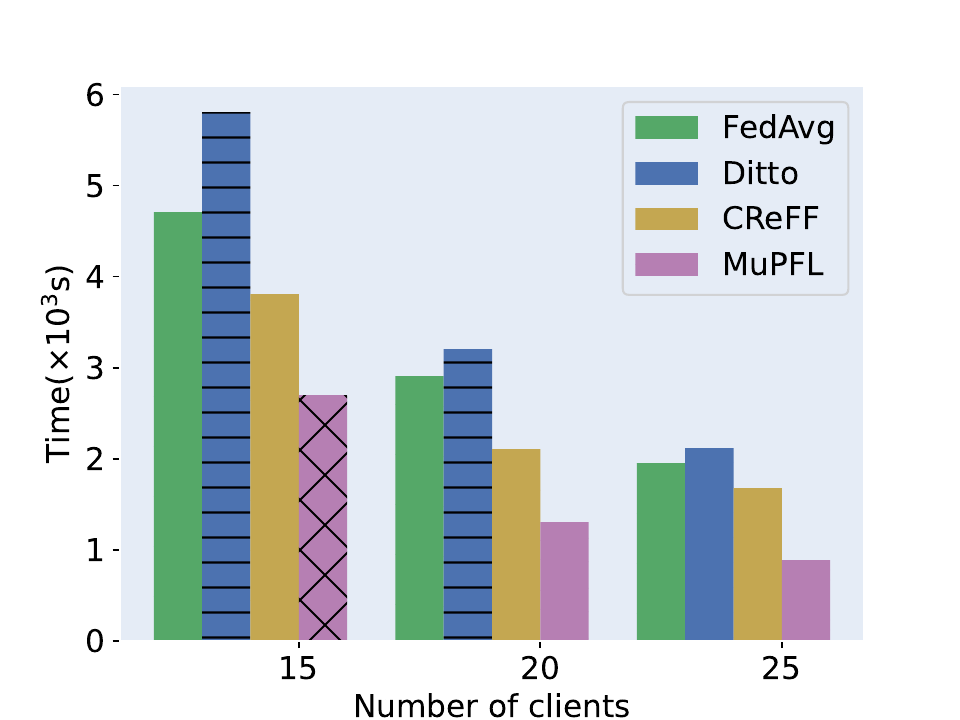}}
\subfigure[Communication rounds to reach target accuracy(RoA@X)]{\includegraphics[width=0.24\textwidth]{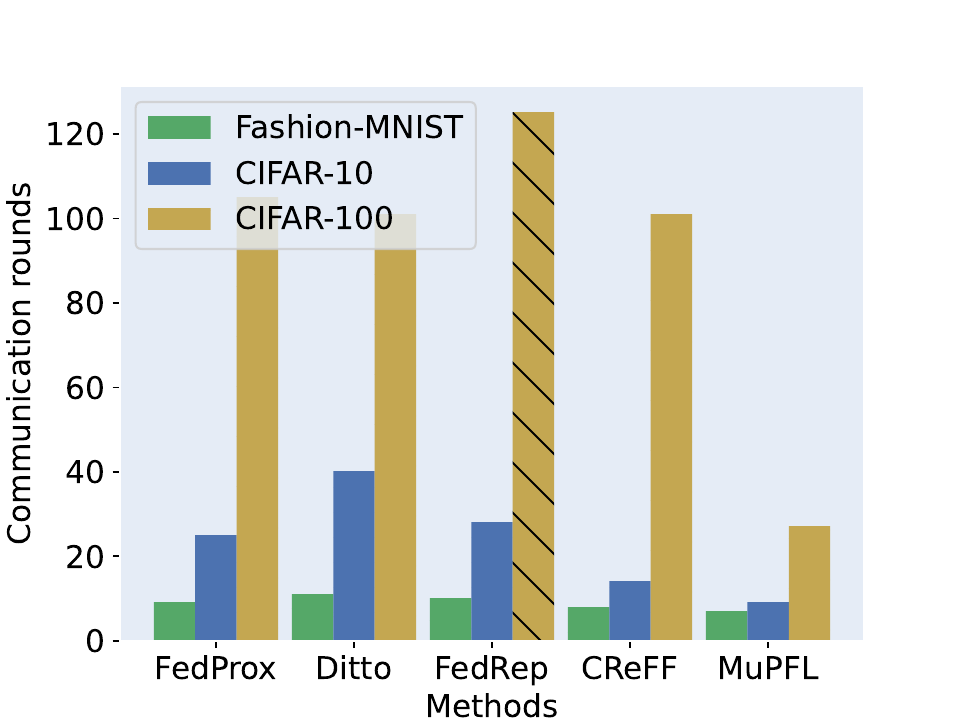}}
\caption{Comparison of training efficiency between MuPFL and other baseline methods in terms of time cost and communication rounds.}
\label{fig:efficiency}
\end{figure}

\setlength\tabcolsep{9pt}%调列距
\begin{table}[t]
	\centering
 \footnotesize
 	\caption{Semantic segmentation results on Cityscapes validation set with different categories ratio IF based on mIoU and mAcc with $\beta=1$.}
	 \resizebox{0.95\columnwidth}{!}{
		\begin{tabular}{ccccc}
  \toprule
            \multirow{2}*{Method}     & \multicolumn{2}{c}{mIoU(\%)} & \multicolumn{2}{c}{mAcc(\%)}   \\
            \cmidrule(r){2-3} \cmidrule(r){4-5}
            \multirow{2}{*}{} & 
            IF=10 & IF=100  & IF=10 & IF=100    \\
            \cmidrule(r){2-2}\cmidrule(r){3-3} \cmidrule(r){4-4} \cmidrule(r){5-5}
            \textsc{FedAvg}  & 44.46 & 45.02 & 87.42 & 87.63 \\ 
            \textsc{FedProx}  & 43.24 & 44.12 & 85.67 & 86.41 \\
            \textsc{FedNova}  & 44.84 & 44.97 & 87.62 & 88.11 \\
            \midrule
            \textsc{Ditto}  & 45.04 & 45.21 & 85.26 & 87.84 \\
		\textsc{FedRep}  & 44.37 & 45.74 & 87.36 & 88.73 \\
            \midrule
            \textsc{CReFF}  & 45.72 & 46.30 & 87.52 & 88.11 \\
            \textsc{MuPFL}  & 47.04 & 48.50 & 88.66 & 89.26 \\
            \bottomrule
		\end{tabular}
  }
	\label{real_results}
\end{table}

\setlength\tabcolsep{4pt}%调列距
\begin{table}[t]
	\centering
 \footnotesize
 	\caption{Analysis of the different number of local epoch $\phi$ and diifferent semantic classes $\beta$ on Cityscapes validation set based on mIoU and mAcc with IF=10.}
	 \resizebox{0.95\columnwidth}{!}{
		\begin{tabular}{ccccccc}
  \toprule
            \multirow{2}*{local epoch}     & \multicolumn{3}{c}{mIoU(\%)} & \multicolumn{3}{c}{mAcc(\%)}   \\
            \cmidrule(r){2-4} \cmidrule(r){5-7}
            \multirow{2}{*}{} & 
            $\beta$=1 & $\beta$=2  & $\beta$=3 & $\beta$=1 & $\beta$=2  & $\beta$=3    \\
            \cmidrule(r){2-2}\cmidrule(r){3-3} \cmidrule(r){4-4} \cmidrule(r){5-5} \cmidrule(r){6-6} \cmidrule(r){7-7}
            $\phi=1$  & 28.84 & 30.05 & 32.44 & 71.77 & 73.74 & 74.93 \\            
            $\phi=5$  & 43.45 & 44.11 & 45.87 & 86.56 & 87.41 & 88.27 \\
            $\phi=10$  & 47.04 & 48.98 & 50.22 & 88.66 & 89.31 & 90.12 \\
            $\phi=20$  & 50.42 & 51.01 & 53.56 & 90.24 & 90.79 & 91.48 \\
            \bottomrule
		\end{tabular}
  }
	\label{local_epoch}
\end{table}

\subsection{Improving training efficiency}
Addressing the quintessential challenge of model personalization in long-tailed data scenarios, where local models often struggle to perform adequately on minority classes, is a key focus of our research. In this vein, we have conducted experiments to demonstrate the strengths of our MuPFL framework in not only personalizing models but also in enhancing training efficiency. We report the time taken for a subset of local clients {15, 20, 25} to reach a benchmark accuracy of 70\% on CIFAR-10, with the findings depicted in~\cref{fig:efficiency}(a). Our results indicate that MuPFL achieves a speedup ranging from 1.35$\times$ to 2.11$\times$ relative to alternative non-i.i.d. and long-tailed approaches, such as Ditto and CReFF, thereby enabling a greater number of clients to attain personalization with satisfactory accuracy.

% To make a more comprehensive study of the acceleration effect of MuPFL, we also conduct experiments to compare the communication rounds each baseline takes to reach the target accuracy in each classification dataset, and the results are shown in ~\cref{fig:efficiency}(b). Specifically, we report communication rounds needed for each benchmark with \textbf{RoA@90} (\textit{i.e., 90\% accuracy}) for Fashion MNIST, \textbf{RoA@70} for CIFAR-10, and \textbf{RoA@35} for CIFAR-100. BAVD can retain the most meaningful features from the local images as much as possible and the generated activation value maps gradually form a personalized signal to guide future training, MuPFL can significantly increase the training speed. Moreover, AMCU and PKCF enhance the model performance, which leads to further acceleration. However, MuPFL does not open a significant gap with other baselines when training with the Fashion-MNIST dataset with only one communication round ahead of CReFF. We can observe that MuPFL can accelerate the training by up to \textbf{90} communication rounds when training CIFAR-100 compared with other baselines, which is over 5.5$\times$ acceleration. The advantages of MuPFL will be more obvious as the complexity of the dataset increases.

To provide a more thorough examination of MuPFL's acceleration capabilities, we have conducted experiments comparing the number of communication rounds required for each baseline to attain target accuracy across different classification datasets. The findings, illustrated in~\cref{acc}(b), indicate that MuPFL necessitates significantly fewer communication rounds for benchmarks with \textbf{RoA@90}(i.e., 90\% accuracy) for Fashion MNIST, \textbf{RoA@70} for CIFAR-10, and \textbf{RoA@35} for CIFAR-100. Despite a higher computational demand per local training epoch compared to baseline methods, MuPFL benefits from BAVD's retention of meaningful local image features and AMCU and PKCF's enhanced model performance, which collectively expedite training. This results in a remarkable reduction of up to 90 communication rounds for CIFAR-100, translating to an acceleration exceeding 5.5$\times$. The efficiency of MuPFL is more pronounced with the increasing complexity of datasets. However, the advantage is less marked with Fashion-MNIST, where it leads by a single round over CReFF. Fundamentally, the total time to reach target accuracy in federated learning is the product of training time, local epoch $\phi$, and communication rounds. With the local epoch $\phi$ set as a constant hyperparameter, MuPFL ensures a substantial decrease in overall training time with much fewer communication rounds required.

\subsection{Ablation study}
Our ablation studies for MuPFL meticulously evaluate the individual and combined effects of BAVD, ACMU, and PKCF on classification tasks. As shown in \cref{abla}, each module independently improves CIFAR-10 accuracy by upwards of 10\%. The combined use of ACMU and BAVD contributes to an additional 4\% increase in CIFAR-100 accuracy by leveraging activation maps to discern crucial feature patterns. The integration of PKCF builds on this foundation, further elevating CIFAR-100 accuracy by more than 6\%, highlighting the compound efficacy of these components in our framework.

The advantages of adaptive clustering over predetermined cluster counts are underscored in \cref{speed}. Training with a static number of clusters leads to subpar results across various models with 90.46\% on FashionMNIST when $\kappa=4$ and $\alpha=0.5$, whereas adaptive clustering surpasses these by margins reaching 2.67\%, validating the approach of customizing cluster sizes to the nuances of local model features for enhanced performance and underscoring the efficacy of ACMU across various classification datasets and different non-i.i.d scenarios.

% \begin{figure}[t]
% \centering
% \subfigure[K-Means-K=5]{\includegraphics[width=0.24\textwidth]{images/kmeans.png}}
% \subfigure[ACMU-$\kappa$=5]{\includegraphics[width=0.24\textwidth]{images/acmu.png}}
% \subfigure[ACMU-Adaptive]{\includegraphics[width=0.24\textwidth]{images/hard.png}}
% \subfigure[MuPFL-Adaptive]{\includegraphics[width=0.24\textwidth]{images/mupfl.png}}
% \caption{PCA visualization of the client parameters with five clusters on FashionMNIST. K-Means-K=5 suggests we use the K-Means algorithm to perform model clustering with five central points. ACMU-$\kappa$=5 and ACMU-Adaptive indicate that we only conduct ACMU in MuPFL with fixed five clustering and adaptively choose the cluster number. MuPFL-Adaptive suggests the complete version of MuPFL.}
% \label{fig:tsne}
% \end{figure}

\begin{table}[t]
  \caption{Different $\kappa$ trained with over the experiments setting $IF=100$ in terms of accuracy on three datasets.}
  \footnotesize
  \centering
  \label{tab:freq}
  	\resizebox{0.95\columnwidth}{!}{%
  \setlength{\tabcolsep}{1mm}{
  \begin{tabular}{ccccccc}
    \toprule
    \multicolumn{1}{c}{\multirow{2}{*}{Method}}& \multicolumn{2}{c}{FashionMNIST} & \multicolumn{2}{c}{CIFAR-10} & \multicolumn{2}{c}{CIFAR-100}\\
    \cmidrule(r){2-3}\cmidrule(r){4-5}\cmidrule(r){6-7}
      & $\alpha$=0.5 & $\alpha$=1.0 & $\alpha$=0.5 & $\alpha$=1.0 & $\alpha$=0.5 & $\alpha$=1.0\\
    \midrule
    $\kappa=$2 & 89.94 & 91.65 & 69.82 & 72.14 & 39.57 & 43.21\\
    $\kappa=$3 & 90.21 & 92.56 & 69.37 & 73.26 & 40.11 & 43.66\\
    $\kappa=$4 & 90.46 & 91.22 & 68.48 & 74.55 & 40.10 & 42.15\\
    $\kappa=$5 & 90.24 & 92.55 & 69.65 & 72.35 & 40.26 & 43.65\\
    $\kappa=$6 & 89.41 & 93.11 & 68.78 & 73.39 & 38.84 & 42.89\\
        \midrule
Adaptive & \textbf{91.90} & \textbf{93.89} & \textbf{70.02} & \textbf{74.76} & \textbf{40.55} & \textbf{43.91}\\
  \bottomrule
\end{tabular}}
}
\label{speed}
% \vspace{-2ex}
\end{table}

\begin{table}[t]
\small
\centering
\setlength{\tabcolsep}{1pt}
\begin{center}      
\caption{Ablation study on three classification datasets Fasion-MNIST, CIFAR-10, and CIFAR-100 with $IF=10, \alpha=0.5$. }
	\resizebox{0.95\columnwidth}{!}{%
 \begin{tabular}{c|ccc|ccc}
\toprule 
Method & BAVD & ACMU & PKCF & F-MNIST & CIFAR-10 & CIFAR-100 \\
\midrule
$Ex_{0}$ & \Checkmark & - & - & 88.46 & 64.87  & 35.67    \\
$Ex_{1}$ & - & \Checkmark & - & 90.23 & 69.98  & 38.33   \\
$Ex_{2}$ & - & - & \Checkmark & 91.45 & 71.41 & 43.56 \\
\midrule
$Ex_{3}$ & \Checkmark & \Checkmark & - & 92.58 & 72.34  & 42.22   \\
$Ex_{4}$ & - & \Checkmark & \Checkmark & 93.24 & 74.88  & 48.02   \\
$Ex_{5}$  & \Checkmark & - & \Checkmark &93.12 & 75.01  & 47.98   \\
\midrule
$Ex_{6}$ & \Checkmark & \Checkmark & \Checkmark & \textbf{94.11} & \textbf{75.88} & \textbf{48.97}   \\
\bottomrule
\end{tabular}
}
\label{abla}
\end{center}
\end{table}

% \vspace{-2ex}
\subsection{Hyper-parameter studies}
We examine the impact of hyper-parameters $m$ (number of features) and $\tau$ (training rounds) in PKCF through a case study on CIFAR-10 using MuPFL with a ResNet-8 architecture across various settings of class distribution skew $\alpha$ and imbalance factor IF. Initially, with $\tau$ fixed at 30, we alter $m$ to ascertain its effect as depicted in ~\cref{fig:hyper}(b). At $m=0$, MuPFL defaults to FedAvg, whereas increasing $m$ proportionally enhances test accuracy, plateauing beyond a certain threshold. Subsequently, we scrutinize the second pivotal hyper-parameter $\tau$, with $m$ set to 20, shown in ~\cref{fig:hyper}(c). The accuracy trend exhibits an inverted "U"-shape, suggesting that while increasing $\tau$ initially ingrains more global knowledge into the local classifier, surpassing approximately 100 training rounds overgeneralizes the classifier, diminishing its local dataset recognition capability.

Our detailed evaluation of MuPFL on the Cityscapes benchmark, summarized in \cref{local_epoch}, demonstrates that increasing the number of local epochs ($\phi$) substantially improves the model's validation performance. Specifically, enhancing $\phi$ from 1 to 20 leads to a significant rise in mIoU and mACC, as evidenced by the validation metrics for different non-i.i.d. data distributions characterized by $\beta=1,2,3$. For instance, when $\phi$ is set to 20, the mIoU exhibits a marked increase from 28.84\% to 50.42\% for $\beta=1$, and a similar upward trend from 32.44\% to 53.56\% for $\beta=3$. Additionally, mAcc follows an analogous trajectory, advancing from 71.77\% to 90.24\% for $\beta=1$, and from 74.93\% to 91.48\% for $\beta=3$. These results underscore the efficacy of BAVD during the local training phase in federated learning environments. Moreover, they emphasize the advantage of the ACMU in handling the challenges posed by non-i.i.d. data distributions.

\begin{figure}[t]
    \centering
    \includegraphics[width=0.48\textwidth]{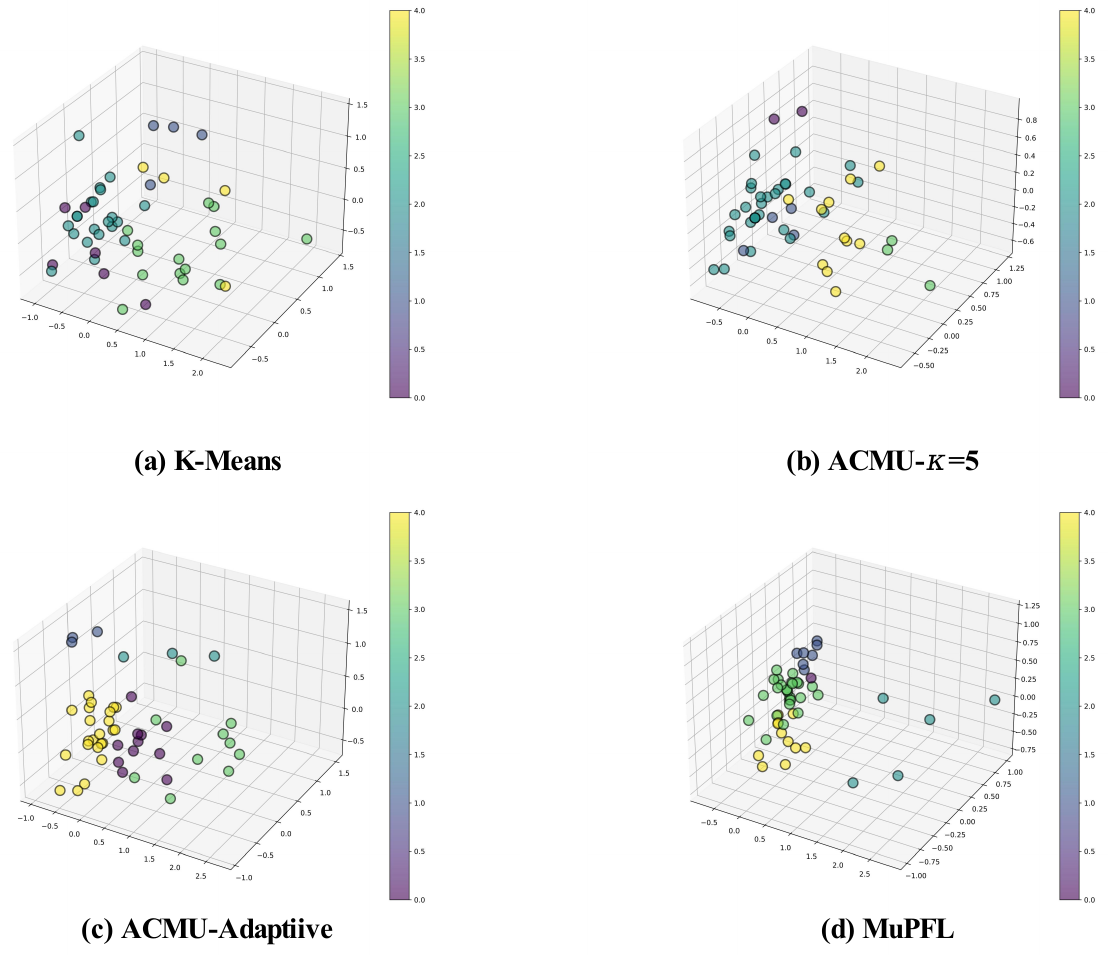}
    \caption{Client parameters visualization with five clusters on FashionMNIST with PCA dimensional reduction. K-Means suggests we use the K-Means algorithm to perform model clustering with five central points. ACMU-$\kappa$=5 and ACMU-Adaptive indicate that we only conduct ACMU in MuPFL with fixed five clustering and adaptively choose the cluster number. MuPFL-Adaptive suggests the complete version of MuPFL.}
    \label{fig:pca}
\end{figure}

\subsection{Visualization and qualitative analysis}
The PCA visualizations in ~\cref{fig:pca} illustrate how client parameter distributions are divided into five clusters. The analysis shows that K-Means clustering leads to less effective grouping of clients, whereas ACMU provides more distinct separation. Adaptive clustering further sharpens these divisions, creating clearer boundaries. This effect is particularly pronounced when ACMU is integrated with BAVD and PKCF, leading to the most effective performance.

The visual comparisons presented in ~\cref{fig:vis} illustrate the qualitative superiority of our MuPFL approach over existing methods. As evidenced in \cref{fig:vis}, MuPFL delivers enhanced detail and contextual fidelity in its outputs, emphasizing its robust performance in real-time semantic segmentation tasks. We can observe that the images inferred by MuPFL exhibit enhanced segmentation capabilities, effectively distinguishing larger foreground objects like taxis and sidewalks, as well as smaller background samples such as pedestrians and street lamps. 
% This is particularly noteworthy under highly imbalanced data conditions, showcasing MuPFL's potential in handling general scene representations effectively.

\section{Conclusion}

In this study, we unveil MuPFL, an advanced framework designed to address the non-i.i.d. and long-tailed distributions in Federated Learning, pivotal for autonomous driving applications. MuPFL sets itself apart by facilitating client-specific localization, thus markedly improving accuracy together with BAVD, ACMU, and PKCF. Our empirical analysis confirms MuPFL's superiority over traditional CNN models in non-i.i.d. contexts, with up to a 7.39\% boost in accuracy. Ablation and hyperparameter studies underscore its efficacy. Prospectively, MuPFL's adaptation to a wider range of non-i.i.d. distributions is anticipated, advancing its integration into the sophisticated milieu of autonomous driving technologies.

% \section{Acknowledgement}
% Portions of this work were presented at the International Conference on Multimedia and Expo. (ICME) in 2023, \textit{Cluster-driven GNN-based Federated Recommendation with Biased Activation Value Dropout}. The work was supported in part by the Basic Research Project No. HZQB-KCZYZ-2021067 of Hetao Shenzhen-HK S\&T Cooperation Zone, by National Natural Science Foundation of China (Grant No. 62102342), by Guangdong Basic and Applied Basic Research Foundation (Grant No. 2023A1515012668), by Shenzhen Science and Technology Program (Grant No. RCBS20221008093120047), by Shenzhen Outstanding Talents Training Fund 202002, by Guangdong Research Projects No. 2017ZT07X152 and No. 2019CX01X104, by the Guangdong Provincial Key Laboratory of Future Networks of Intelligence (Grant No. 2022B1212010001), and by The Major Key Project of PCL Department of Broadband Communication.

% \normalem
\bibliographystyle{IEEEtran}
\bibliography{mybibfile}

\vfill

\end{document}